\documentclass[10pt,letterpaper,twocolumn]{article}
\usepackage[T1]{fontenc}
\usepackage[utf8]{inputenc}
\usepackage[letterpaper,left=0.75in,right=0.75in,top=0.65in,bottom=0.70in,footskip=25pt,columnsep=0.25in]{geometry}
\usepackage{newtxtext}
\usepackage[table]{xcolor}
\usepackage{graphicx,amsmath,amssymb}
\usepackage{booktabs,array,tabularx,makecell,multirow}
\usepackage{caption}
\usepackage[caption=false]{subfig}
\usepackage[round,authoryear]{natbib}
\let\cite\citep
\setcitestyle{aysep={}}
\usepackage[hyphens]{url}
\usepackage{microtype,tikz,etoolbox,needspace}
\usepackage{hyperref}
\usepackage{bookmark}
\definecolor{OceanBlue}{HTML}{0077B6}
\colorlet{OceanLink}{OceanBlue}
\definecolor{OceanHeading}{HTML}{0D3B66}
\hypersetup{
  colorlinks=true,linkcolor=OceanLink,citecolor=OceanLink,urlcolor=OceanLink,
  breaklinks=true,
  pdftitle={One Ocean, Many Tasks: A Data-Grounded Simulation Suite and Benchmark for Marine Robotics},
  pdfauthor={Shuaijun Liu, Qifu Wen, Xingwei Chen, Xinwen Hu, Feiyang You, Sijia Chen, Ningxin Su},
  pdfsubject={Data-grounded simulation and benchmarking for marine robotics},
  pdfkeywords={marine robotics, ocean simulation, benchmark, multi-agent systems}}
\newcommand{\revise}[1]{#1}
\definecolor{oursbg}{RGB}{255,250,230}
\definecolor{scenarioGray}{RGB}{245,245,245}
\definecolor{bestblue}{RGB}{0,56,128}
\newcommand{\best}[1]{\textcolor{bestblue}{\textbf{#1}}}
\definecolor{symgreen}{RGB}{140,200,140}
\definecolor{symred}{RGB}{235,170,170}
\newcommand{\symdot}[1]{\textcolor{#1}{\raisebox{-0.45\height}{\scalebox{1.12}{\Large\textbullet}}}}
\newcommand{\symyes}{\symdot{symgreen}}
\newcommand{\symno}{\symdot{symred}}

\newcolumntype{Y}{>{\centering\arraybackslash}X}
\makeatletter
\renewcommand{\section}{\@startsection{section}{1}{\z@}%
  {-2.5ex \@plus -1ex \@minus -.2ex}{1.1ex \@plus .2ex}%
  {\normalfont\large\bfseries\color{OceanHeading}}}
\renewcommand{\subsection}{\@startsection{subsection}{2}{\z@}%
  {-2.2ex \@plus -.8ex \@minus -.2ex}{.8ex \@plus .2ex}%
  {\normalfont\normalsize\bfseries\color{OceanHeading}}}
\pretocmd{\section}{\Needspace{4\baselineskip}}{}{}
\makeatother

\newcommand{\OceanWebsiteIcon}{\makebox[12pt][c]{\raisebox{-1.5pt}{%
  \includegraphics[height=10.2pt]{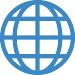}}}}
\newcommand{\OceanGitHubIcon}{\makebox[12pt][c]{\raisebox{-1.6pt}{%
  \includegraphics[height=10.4pt]{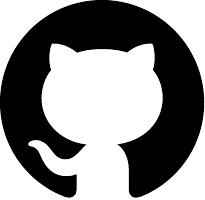}}}}
\newcommand{\OceanHFIcon}{\makebox[11pt][c]{\raisebox{-1.4pt}{%
  \includegraphics[width=11pt]{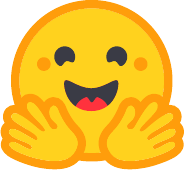}}}}
\let\OceanDemoIcon\OceanWebsiteIcon
\newcommand{\OceanResourceLinks}{%
  \par\vspace{16pt}\noindent
  \OceanWebsiteIcon\,\href{https://nebulis-lab.github.io/OneOcean/}{\textbf{Project}}\hfill
  \OceanGitHubIcon\,\href{https://github.com/NEBULIS-Lab/OneOcean}{\textbf{Code}}\hfill
  \OceanDemoIcon\,\href{https://nebulis-lab.github.io/OneOcean/demo/}{\textbf{Demo}}\hfill
  \OceanHFIcon\,\href{https://huggingface.co/datasets/anonymous321123/OneOcean_Environment_Dataset}{\textbf{Hugging Face}}\par}
\renewcommand{\maketitle}{%
  \begingroup\parindent=0pt\centering
  {\color{black}\hrule height .8pt}\vspace{8pt}
  {\normalfont\fontsize{18}{21}\selectfont\bfseries
    One Ocean, Many Tasks: A Data-Grounded\\
    Simulation Suite and Benchmark for Marine Robotics\par}
  \vspace{8pt}{\color{black}\hrule height .35pt}\vspace{9pt}
  {\normalfont\normalsize\bfseries
    Shuaijun Liu\textsuperscript{1}, Qifu Wen\textsuperscript{2,3},
    Xingwei Chen\textsuperscript{1}, Xinwen Hu\textsuperscript{4},
    Feiyang You\textsuperscript{1}, Sijia Chen\textsuperscript{1},
    Ningxin Su\textsuperscript{1,*}\par}
  \vspace{4pt}{\normalfont\small
    \textsuperscript{1}The Hong Kong University of Science and Technology (Guangzhou)\quad
    \textsuperscript{2}Boston University\\[2pt]
    \textsuperscript{3}Shanghai Jiao Tong University\quad
    \textsuperscript{4}Tencent\quad
    \textsuperscript{*}Corresponding author:
    \href{mailto:ningxinsu@hkust-gz.edu.cn}{ningxinsu@hkust-gz.edu.cn}\par}
  \endgroup}
\newsavebox{\OceanAbstractBox}
\renewenvironment{abstract}{%
  \par\noindent\begin{lrbox}{\OceanAbstractBox}%
  \begin{minipage}{\dimexpr\columnwidth-16.7pt\relax}%
  \normalfont\small
  {\centering\normalsize\bfseries\color{OceanHeading}Abstract\par}\vspace{4pt}%
}{%
  \OceanResourceLinks
  \par\end{minipage}\end{lrbox}%
  \noindent\begin{tikzpicture}
    \node[draw=OceanBlue!35,fill=OceanBlue!4,line width=.35pt,
      rounded corners=3pt,inner sep=8pt]{\usebox{\OceanAbstractBox}};
  \end{tikzpicture}\par\vspace{4pt}}

\begin{document}
\twocolumn[{%
  \begin{minipage}{\textwidth}
    \maketitle
    \vspace{8pt}
    \centering
    \captionsetup{hypcap=false}
    \includegraphics[width=\textwidth]{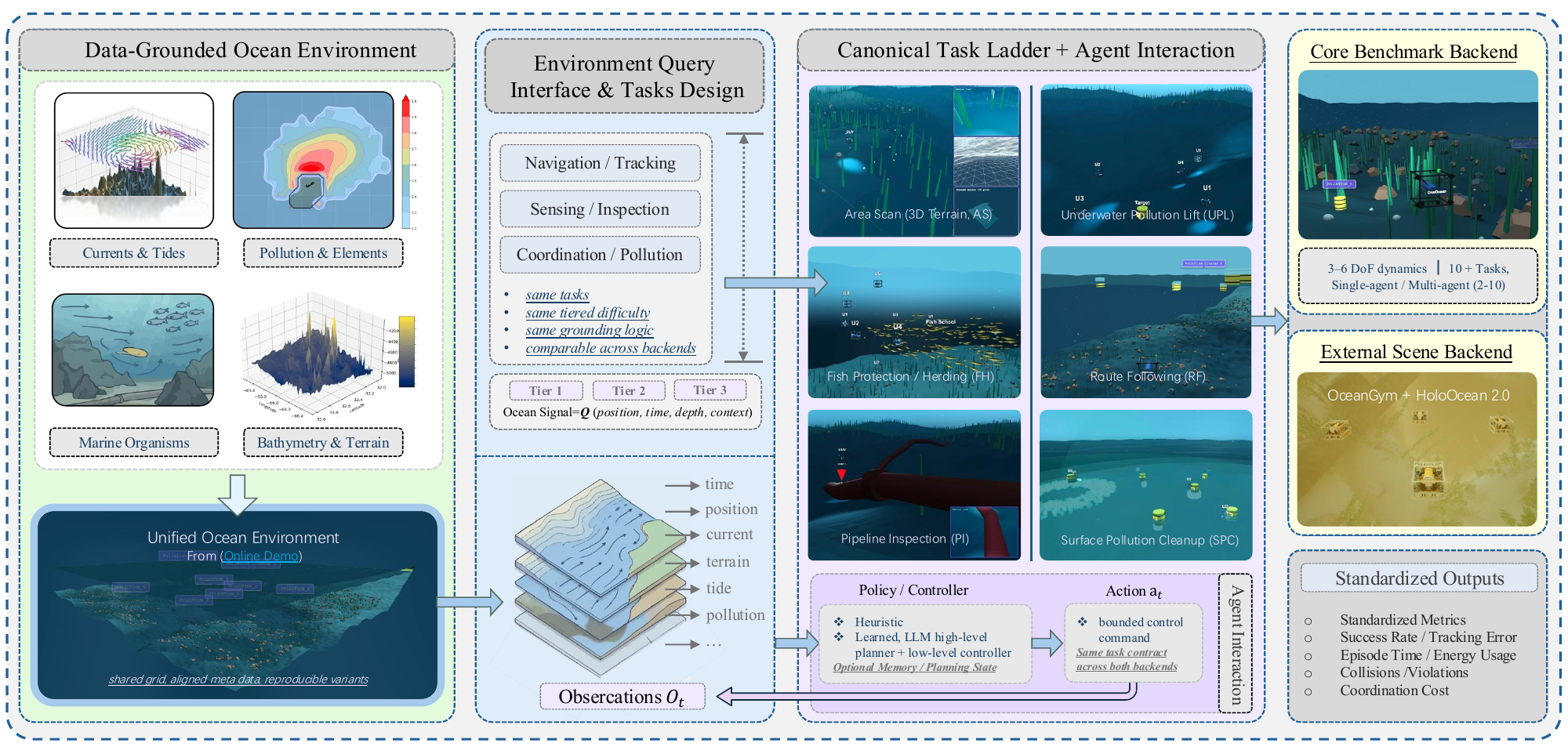}
    \captionof{figure}{OneOcean overview. A unified ocean-environment product grounds a canonical multi-task benchmark and two complementary execution backends, enabling reproducible evaluation across navigation, sensing, coordination, and pollution-related marine-robotics tasks.}
    \label{fig:overview}
    \vspace{10pt}
  \end{minipage}%
}]
\begin{abstract}
Simulation is essential for marine robotics, yet existing platforms often trade off between visually realistic worlds, high-throughput experimentation, and ocean-condition realism grounded in real data. We present OneOcean, a data-grounded simulation suite and benchmark built around a unified spatiotemporal ocean-environment product that harmonizes bathymetry with data-driven currents and tides, with optional pollution fields for cleanup-related evaluation. On top of this environment representation, OneOcean defines a task ladder spanning navigation and station-keeping under currents, waypoint and route following, depth-profile tracking, area scanning, pipeline inspection, and multi-agent coordination, including formation transit, fish protection and patrol (herding), and surface pollution localization, containment, and cleanup. The suite supports multi-agent settings with 2--10 vehicles. Our simulator supports three to six Degrees of Freedom (DoF) and produces standardized metrics and reproducible run manifests, enabling consistent aggregation across tasks, difficulties, and scaling settings. Experiments across dataset variants and scenes systematically stress planning and control under data-grounded disturbances and reveal trade-offs among heuristic baselines, Behavior Cloning (BC), and a Large Language Model (LLM)-based planner.

\end{abstract}
\section{Introduction}

\revise{Marine robots must operate under complex, highly variable ocean conditions. In practice, \textbf{currents and tides} can dominate vehicle motion, while field trials remain expensive and time-limited and at-sea sensing and communication are constrained \cite{Trends_Marine_Robotics}. Simulation is therefore essential, but existing platforms often trade off between (i) visually realistic underwater worlds, (ii) high-throughput, reproducible experimentation, and (iii) \textbf{ocean-condition realism grounded in real data}.}

\revise{We present \textbf{OneOcean}, a data-grounded simulation suite and benchmark for marine robotics. OneOcean is built around a unified spatiotemporal ocean-environment dataset in Network Common Data Form (NetCDF). It harmonizes bathymetry from the General Bathymetric Chart of the Oceans (GEBCO) \cite{GEBCO2025} with data-driven currents and tides from the Copernicus Marine Service Global Ocean Physics Analysis and Forecast (GOPAF) product \cite{copernicus,CopernicusGOPAF2025}. On top of this environment representation, OneOcean defines a \textbf{10-task ladder} spanning single-agent navigation and tracking, spatial sensing and coverage, and multi-agent coordination and pollution-related tasks, with support for teams of 2--10 vehicles. The same environment product is also packaged into multiple dataset variants so the benchmark can support fast local iteration, scene-oriented construction, and broader public release without changing task semantics. Figure~\ref{fig:overview} summarizes this organization.}

\begin{table*}[!t]
\centering
\scriptsize
\setlength{\tabcolsep}{0.6pt}
\renewcommand{\arraystretch}{1.02}
\textbf{Symbols:} \symyes\ Provided; \symno\ Not provided. \quad
\textbf{Tasks:} G2G, SK, RF, DPT, AS, PI, FT, SPC, UPL, and FH.\par
\begin{tabularx}{\textwidth}{@{}l c | *{9}{Y} *{10}{c}@{}}
\toprule
\textbf{Platform or Work} & \textbf{Venue}
& \makecell{\textbf{Real-data}\\ \textbf{In Loop}}
& \makecell{\textbf{Multi-}\\\textbf{agent}}
& \makecell{\textbf{Cross-}\\\textbf{backend}}
& \makecell{\textbf{3D}\\\textbf{Currents}}
& \makecell{\textbf{Pollution}\\\textbf{or Spill}}
& \makecell{\textbf{Tide}\\\textbf{Modeling}}
& \makecell{\textbf{Canonical}\\\textbf{Task}}
& \makecell{\textbf{Metrics}\\\textbf{Protocol}}
& \makecell{\textbf{Headless}\\\textbf{Batch}}
& \textbf{G2G} & \textbf{SK} & \textbf{RF} & \textbf{DPT} & \textbf{AS} & \textbf{PI} & \textbf{FT} & \textbf{SPC} & \textbf{UPL} & \textbf{FH} \\
\midrule

MIMIR-UW & 2023
& \symno & \symno & \symno
& \symno & \symno & \symno & \symno & \symno
& \symyes
& \symno & \symno & \symno & \symno & \symno & \symno & \symno & \symno & \symno & \symno \\

HoloOcean 2.0 & 2025
& \symno & \symyes & \symno
& \symyes & \symno & \symno & \symno & \symno
& \symyes
& \symyes & \symyes & \symyes & \symyes & \symyes & \symyes & \symyes & \symno & \symno & \symno \\

OceanSim & 2025
& \symno & \symyes & \symno
& \symno & \symno & \symno & \symno & \symno
& \symyes
& \symyes & \symno & \symyes & \symno & \symno & \symyes & \symyes & \symno & \symno & \symno \\

MarineGym & IROS 2025
& \symno & \symyes & \symyes
& \symyes & \symyes & \symyes & \symyes & \symyes
& \symyes
& \symyes & \symyes & \symyes & \symno & \symno & \symno & \symyes & \symno & \symno & \symno \\

OceanGym & 2025
& \symno & \symyes & \symyes
& \symyes & \symyes & \symno & \symyes & \symyes
& \symyes
& \symyes & \symyes & \symyes & \symno & \symyes & \symno & \symyes & \symno & \symno & \symno \\

URLab & ICRA 2026
& \symno & \symno & \symyes
& \symno & \symno & \symno & \symno & \symyes
& \symno
& \symyes & \symno & \symno & \symno & \symno & \symno & \symno & \symno & \symno & \symno \\

DREAM & ICRA 2026
& \symno & \symno & \symno
& \symno & \symno & \symno & \symno & \symyes
& \symno
& \symno & \symno & \symno & \symno & \symyes & \symyes & \symno & \symno & \symno & \symno \\

Douro MARL & ICRA 2026
& \symyes & \symyes & \symno
& \symno & \symyes & \symyes & \symno & \symyes
& \symyes
& \symno & \symno & \symno & \symno & \symyes & \symno & \symno & \symno & \symno & \symno \\

Map-TIDAL & ICRA 2026
& \symno & \symyes & \symno
& \symno & \symno & \symno & \symno & \symyes
& \symno
& \symno & \symno & \symno & \symno & \symyes & \symyes & \symno & \symno & \symno & \symno \\

CAMAR & AAAI 2026
& \symno & \symyes & \symno
& \symno & \symno & \symno & \symyes & \symyes
& \symyes
& \symyes & \symno & \symno & \symno & \symno & \symno & \symno & \symno & \symno & \symno \\

RobotArena $\infty$ & ICLR 2026
& \symyes & \symno & \symno
& \symno & \symno & \symno & \symyes & \symyes
& \symyes
& \symno & \symno & \symno & \symno & \symno & \symno & \symno & \symno & \symno & \symno \\

CaP-X & ICML 2026
& \symno & \symno & \symyes
& \symno & \symno & \symno & \symyes & \symyes
& \symyes
& \symno & \symno & \symno & \symno & \symno & \symno & \symno & \symno & \symno & \symno \\

\rowcolor{gray!20}%
\textbf{OneOcean (ours)} & 2026
& \symyes & \symyes & \symyes
& \symyes & \symyes & \symyes & \symyes & \symyes
& \symyes
& \symyes & \symyes & \symyes & \symyes & \symyes & \symyes & \symyes & \symyes & \symyes & \symyes \\

\bottomrule
\end{tabularx}
\caption{System-level comparison and task coverage. Entries are conservatively coded from the cited papers and their public artifacts \cite{MIMIRUW,HoloOcean2,OceanSim,MarineGym,OceanGym,URLab2026,DREAM2026,DouroMARL2026,MapTIDAL2026,CAMAR2026,RobotArena2026,CaPX2026}.}
\label{tab:platform_features_tasks}
\end{table*}

\revise{OneOcean provides two complementary execution backends that share task definitions and evaluation semantics. The \emph{core benchmark backend} is deterministic and throughput-first, producing standardized metrics and reproducible run manifests for consistent aggregation. The \emph{external underwater-scene backend} reuses packaged scenes from an existing simulator stack and injects OneOcean's data-grounded currents, enabling qualitative evaluation in visually realistic underwater settings while preserving the benchmark protocol. This split separates controlled large-scale benchmarking from scene-rich portability studies while grounding both backends in the same ocean data product.}
In summary, our contributions are:
\begin{itemize}
    \item \revise{A unified, reusable ocean-environment data product and tooling for exporting data-grounded disturbance signals to simulators, with variants that support fast iteration, scene construction, and broader public release.}
    \item \revise{A canonical 10-task benchmark suite with difficulty levels, standardized metrics, auditable run manifests, and support for 3--6 DoF vehicle dynamics, with 6DoF used for the main quantitative evaluation.}
    \item \revise{Experiments across the two backends that reveal task difficulty, scaling, and planner-compute trade-offs under data-grounded disturbances.}
\end{itemize}

\section{Related Work}
\label{sec:related_work}

Classical simulators built on Gazebo \cite{Gazebo}, including UUV Simulator \cite{UUVSim} and UWSim \cite{UWSim}, provide dynamics and sensor stacks commonly used in robotics workflows, but they are not designed for data-grounded ocean-condition benchmarking at scale. More recent systems target higher-fidelity rendering or higher-throughput experimentation, including MARUS \cite{MARUS}, OceanSim on Isaac Sim \cite{OceanSim,IsaacSim}, DAVE \cite{DAVE}, MarineGym \cite{MarineGym}, and URLab, which integrates Unreal Engine with MuJoCo \cite{URLab2026}. \revise{These platforms provide important simulator infrastructure, but they typically do not expose a reusable ocean-condition product, a fixed task ladder, and a reporting protocol that remain consistent across multiple execution backends. Reproducing the same ocean-conditioned task across simulators therefore requires substantial custom engineering. Table~\ref{tab:platform_features_tasks} summarizes this distinction at both the system and task levels.}

Ocean-condition realism is commonly provided by numerical models and operational forecasting systems. Representative community models include Delft3D \cite{delft3d}, NEMO \cite{nemo}, MITgcm \cite{MITgcm}, ROMS \cite{roms}, and HYCOM \cite{hycom}. Operational products from the Copernicus Marine Service, including GOPAF, provide data-assimilative currents and hydrographic variables. GEBCO provides global gridded seafloor elevation, while NetCDF stores labeled multidimensional coordinates, variables, units, and provenance in a self-describing representation. These sources are public, globally scoped, and machine-queryable, allowing the same ocean condition to be extracted and audited across runs. Toolkits such as Parcels \cite{parcels} and OpenDrift \cite{opendrift} support Lagrangian drift and dispersion analysis. \revise{OneOcean complements these efforts by treating ocean conditions as a reusable spatiotemporal product for robotics and pairing them with a protocol that runs across multiple simulator backends.}

\revise{Recent task-specific marine systems provide complementary evidence. DREAM studies vision-language monitoring through simulation and tank trials \cite{DREAM2026}; Map-TIDAL allocates cooperative underwater-inspection tasks \cite{MapTIDAL2026}; and Douro MARL maps a data-driven river plume with multiple vehicles \cite{DouroMARL2026}. Recent AI benchmarks broaden the evaluation perspective: CAMAR defines high-throughput continuous multi-agent routing and a three-tier protocol \cite{CAMAR2026}; RobotArena $\infty$ converts real demonstrations into scalable simulation tests for robot policies \cite{RobotArena2026}; and CaP-X evaluates executable embodied agents across multiple manipulation suites \cite{CaPX2026}. OneOcean addresses the remaining marine-specific gap by combining standardized environmental conditions, a canonical task suite, auditable run manifests, and shared contracts across two execution backends.}

\section{OneOcean Suite}
\label{sec:oneocean_suite}

OneOcean is organized around a single principle: treat \emph{ocean conditions} as a reusable, auditable product, and evaluate robotics methods on a canonical task ladder under these conditions. This section summarizes the environment product, how it is consumed by the benchmark backends, and the task and evaluation definitions.

\subsection{Unified spatiotemporal ocean-environment product}
\label{sec:env_product}

The shared NetCDF representation harmonizes GEBCO bathymetry with GOPAF currents and optional tides. It exposes named latitude, longitude, time, and depth coordinates, eastward and northward current components, elevation, and a mask for invalid terrain and NoData cells. The pipeline crops the requested region, retains the returned time and depth levels, aligns variables on one horizontal grid, and records provenance.

The core backend selects a fixed $(t,d)$ slice for deterministic rollouts, whereas the external backend advances a dataset clock. Table~\ref{tab:data_variants} reports the realized products: the 0--200\,m requests returned 26 levels through 186.13\,m, while \texttt{tiny} contains one near-surface level at 0.494\,m. Appendix Sec.~\ref{app:data} specifies the full schema, mask semantics, coordinate conventions, validation results, and tide and depth caveats.

The mask denotes invalid terrain, crop-exterior, or NoData cells rather than a guaranteed coastline classification, and consumers must not interpret zero-filled masked elevation as seabed. Likewise, the single \texttt{tiny} layer is not evidence of resolved 3D flow. In the tide-enabled \texttt{scene} product, higher-frequency tide components are aligned to the base clock and attenuated with depth; this documented engineering profile supports controlled disturbance tests but not phase-accurate 3D tidal simulation.

\begin{table}[!t]
\centering
\scriptsize
\setlength{\tabcolsep}{2.2pt}
\renewcommand{\arraystretch}{1.08}
\begin{tabularx}{\columnwidth}{@{}l>{\raggedright\arraybackslash}p{0.30\columnwidth}>{\centering\arraybackslash}p{0.24\columnwidth}>{\raggedright\arraybackslash}X@{}}
\toprule
\textbf{Variant} & \textbf{Region and period} & \textbf{$T\!\times\!D\!\times\!H\!\times\!W$} & \textbf{Resolution; size; use} \\
\midrule
\texttt{tiny} & 42.1--42.7 N, 71.2--70.2 W; 2025 & $357\!\times\!1\!\times\!62\!\times\!101$ & $\approx0.01^\circ$; 29 MB; tests and web \\
\texttt{scene} & 32--33 N, 66.5--65.5 W; Dec. 2025 & $23\!\times\!26\!\times\!240\!\times\!240$ & $0.00418^\circ$; 690 MB; scenes \\
\texttt{public} & 30--40 N, 72--62 W; 2025 & $357\!\times\!26\!\times\!41\!\times\!41$ & $0.25^\circ$; 175 MB; release \\
\bottomrule
\end{tabularx}
\caption{Realized combined NetCDF variants. Sizes are the merged products themselves; source subsets and derived assets are distributed separately.}
\label{tab:data_variants}
\end{table}

\subsection{Data-grounded disturbance signals}
\label{sec:signals}

\begin{figure}[t]
 \centering
 \includegraphics[width=0.98\linewidth]{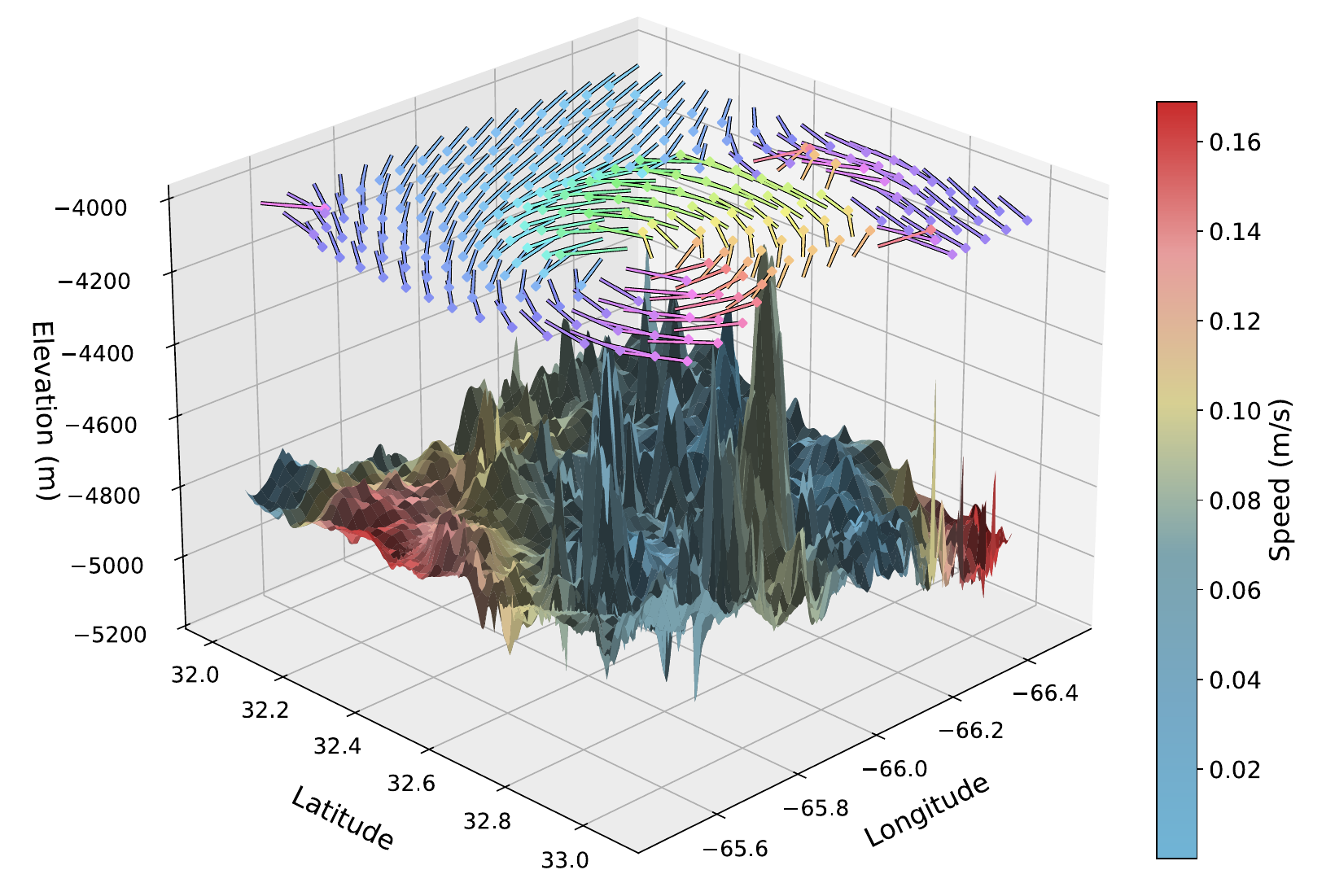}
 \caption{Base current field visualization over bathymetry. The horizontal current components are $u_{\mathrm{data}}$ and $v_{\mathrm{data}}$; surface color indicates speed magnitude (m\,s$^{-1}$). View configuration: pitch $25^\circ$, azimuth $45^\circ$, and depth index $0$\,m. The terrain surface is colored by speed and overlaid with arrows indicating current direction.}
 \label{fig:base_current_field}
\end{figure}

OneOcean exports either a fixed gridded slice for high-throughput rollouts or a temporal profile for scene-backed execution. Both apply current through a recorded relative-velocity convention in the core frame $(x,y,z)=$(east, depth-down, north), and every artifact records its extraction indices. Figure~\ref{fig:base_current_field} visualizes one grounded current slice over bathymetry; Appendix Sec.~\ref{app:data} details the export and provenance contract.

\paragraph{Environment queries and probes.}
\label{sec:queries_probes}

Shared queries expose local current, bathymetry, mask validity, and task signals. Pollution tasks observe $o_i^c(t)=C(p_i(t),t)$ from either a current-advected Gaussian or a regeneratable Ocean Current Prediction Network (OCPNet) field; these are controlled signals, not in-situ labels. Appendix Sec.~\ref{app:models_ros} gives the probe, sink, and discrete-source cleanup semantics.

\subsection{Task ladder and difficulty levels}
\label{sec:tasks}

OneOcean defines ten tasks spanning navigation and tracking (G2G, SK, RF, DPT), sensing and inspection (AS, PI), and coordination and pollution response (FT, SPC, UPL, FH). Table~\ref{tab:task_ladder} gives team sizes and primary metrics; Appendix Sec.~\ref{app:tasks} provides the complete objective, observation, constraint, and success contracts.

\begin{table}[!t]
\centering
\scriptsize
\setlength{\tabcolsep}{3pt}
\renewcommand{\arraystretch}{1.08}
\begin{tabular*}{\columnwidth}{@{\extracolsep{\fill}}>{\raggedright\arraybackslash}p{0.18\columnwidth}c c c >{\raggedright\arraybackslash}p{0.28\columnwidth}@{}}
\toprule
\textbf{Category} & \textbf{Task} & \textbf{Agents} & \textbf{Goal} & \textbf{Primary metric} \\
\midrule
Navigation & \textbf{G2G} & 1 & reach & final goal error \\
Navigation & \textbf{SK} & 1 & hold & station root-mean-square error \\
Navigation & \textbf{RF} & 1 & follow & cross-track error \\
Navigation & \textbf{DPT} & 1 & track & depth-tracking error \\
\makecell[l]{Sensing and\\Coverage} & \textbf{AS} & 1--10 & scan & coverage fraction \\
\makecell[l]{Sensing and\\Inspection} & \textbf{PI} & 1--10 & inspect & inspection score \\
Coordination & \textbf{FT} & 2--10 & transit & formation error \\
Pollution & \textbf{SPC} & 2--10 & clean & cleanup rate \\
Coordination & \textbf{UPL} & 5 & lift & lift success \\
Coordination & \textbf{FH} & 8 & herd & containment success \\
\bottomrule
\end{tabular*}
\caption{Canonical task ladder used in OneOcean (10 tasks). Each task is evaluated at Tiers 1--3 and paired with a compact goal type and a primary task-specific metric.}
\label{tab:task_ladder}
\end{table}

\revise{Tiers 1, 2, and 3 jointly increase current strength, precision, route and coverage extent, sensing difficulty, and coordination tightness while preserving each task's interface. Every task declares its objective, observations, feasibility constraints, success condition, and primary metric. The release also includes a lightweight Robot Operating System 2 (ROS~2) bridge for state, probes, and velocity commands; its frame, topic, and service contract appears in Appendix Sec.~\ref{app:models_ros}, while ROS performance remains outside this evaluation.}

Navigation and tracking tasks test disturbance rejection and sustained trajectory accuracy; sensing and inspection tasks require spatial coverage or physical cue encounters; coordination and pollution tasks add team geometry, assignment, and dwell-time coupling. This grouping keeps the shared reporting contract compact without collapsing task-specific progress into success alone.

\revise{
\begin{table*}[!t]
\centering
\scriptsize
\setlength{\tabcolsep}{1.2pt}
\renewcommand{\arraystretch}{1.10}
\begin{tabularx}{\textwidth}{@{}l*{3}{Y}|*{3}{Y}|*{6}{Y}@{}}
\toprule
\multirow{2}{*}{\textbf{Method}} & \multicolumn{3}{c}{\textbf{Macro (Tier 2)}} & \multicolumn{3}{c}{\textbf{Macro (Tier 3)}} & \multicolumn{6}{c}{\textbf{Tier 3 key-task metrics}} \\
\cmidrule(lr){2-4}\cmidrule(lr){5-7}\cmidrule(lr){8-13}
 & \makecell{Suite\\SR(\%)} & \textbf{$E$} & \makecell{Coll.\\(\%)} & \makecell{Suite\\SR(\%)} & \textbf{$E$} & \makecell{Coll.\\(\%)} & \makecell{G2G\\SR(\%)} & \makecell{RF\\SR(\%)} & \makecell{SK\\RMS} & \makecell{Scan\\cov.} & \makecell{Pipe\\score} & \makecell{Cleanup\\rate} \\
\midrule
\rowcolor{oursbg}
Heuristic & \best{84.6} & 1842.3 & \best{5.68} & \best{29.0} & 3352.4 & 8.19 & \best{74.0} & \best{54.0} & \best{1.07} & \best{0.89} & \best{0.78} & \best{0.56} \\
BC & 58.6 & \best{1271.2} & 9.72 & 6.0 & \best{1635.1} & \best{6.41} & 34.0 & 0.0 & 26.52 & 0.89 & 0.67 & 0.06 \\
\bottomrule
\end{tabularx}
\caption{Headline core-benchmark results (6DoF): macro averages at Tiers 2 and 3 plus Tier 3 key-task metrics, using 50 episodes per task and tier. Success Rate (SR) is reported in percent; Macro Suite SR is an unweighted mean across the 10 canonical tasks. Coll is the mean collision-step ratio (\%) with a 2.0\,m radius. Best values per column are highlighted.}
\label{tab:headline_macro_keymetrics}
\end{table*}
}

\revise{
\begin{table*}[t]
\centering
\scriptsize
\setlength{\tabcolsep}{1.5pt}
\renewcommand{\arraystretch}{1.10}
\begin{tabularx}{\textwidth}{@{}l*{9}{Y}@{}}
\toprule
\textbf{Method} & \makecell{Avg.\\SR(\%)} & \makecell{Cleanup\\SR(\%)} & \makecell{Cleanup\\rate} & \makecell{Scan\\SR(\%)} & \makecell{Scan\\cov.} & \makecell{Pipe\\SR(\%)} & \makecell{Pipe\\score} & \textbf{$T$(s)} & \textbf{$E$} \\
\midrule
\rowcolor{oursbg}
Heuristic & \best{65.8} & 60.0 & \best{0.87} & \best{77.5} & 0.80 & \best{60.0} & \best{0.69} & \best{776.8} & 9132.4 \\
BC & 45.0 & 10.0 & 0.32 & 67.5 & 0.79 & 57.5 & 0.61 & 921.6 & \best{6908.2} \\
LLM-ChatGLM3-6B & 55.0 & 60.0 & \best{0.87} & 77.5 & \best{0.82} & 27.5 & 0.54 & 860.3 & 9671.5 \\
LLM-Qwen2.5-7B & 47.5 & 57.5 & 0.81 & 57.5 & 0.78 & 27.5 & 0.57 & 1077.2 & 11703.8 \\
LLM-Qwen2-7B & 45.0 & 57.5 & 0.86 & 67.5 & \best{0.81} & 10.0 & 0.51 & 991.4 & 10422.1 \\
LLM-Mistral-7B & 42.5 & 55.0 & 0.82 & 55.0 & 0.78 & 17.5 & 0.52 & 1044.2 & 11288.9 \\
LLM-Llama3-8B & 44.2 & \best{67.5} & 0.84 & 47.5 & 0.77 & 17.5 & 0.59 & 1099.1 & 11614.2 \\
LLM-Llama2-7B & 30.8 & 55.0 & \best{0.87} & 27.5 & 0.75 & 10.0 & 0.46 & 1156.4 & 11654.3 \\
\bottomrule
\end{tabularx}
\caption{Core-benchmark planning-suite performance at Tier 2 with doubled currents: heuristic vs BC vs local LLM planners on three planning-sensitive tasks, using 40 episodes per task (120 per method). Best values per column are highlighted.}
\label{tab:planning_suite_perf_t2}
\end{table*}
}

\subsection{Two execution backends}
\label{sec:backends}

OneOcean provides two complementary execution backends that share task and reporting contracts. The \textbf{core benchmark backend} is headless and throughput-first: it consumes a NumPy Archive (NPZ) slice plus task and action configuration and returns observations, metrics, manifests, and recorder streams at scale. It supports 3--6 DoF engineering dynamics (6DoF in the main experiments) \cite{UUVSim}. The \textbf{external underwater-scene backend} prioritizes scene reuse: it consumes a temporal current export and the same task semantics inside packaged underwater scenes, returning quantitative route and task outputs with curated media. Thus, the second backend studies portability and scene realization; it is not claimed as a new rendering engine or as an architecture-matched throughput comparison.

\paragraph{Benchmark protocol and metrics.}
\label{sec:protocol}

\begin{figure}[!b]
 \centering
 \includegraphics[width=0.98\linewidth]{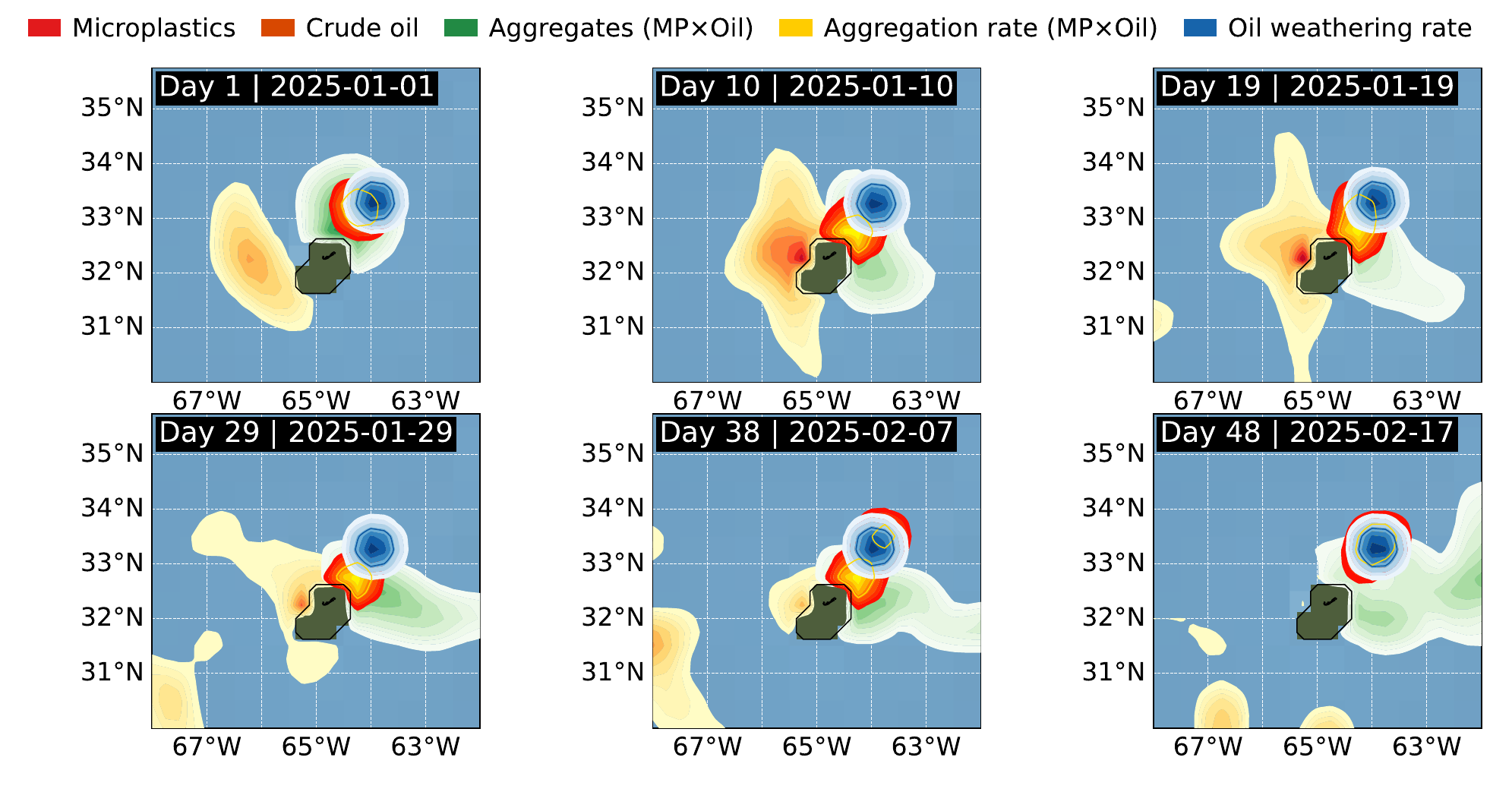}
 \caption{Curated OCPNet multi-species diffusion and reaction field used as qualitative evidence. We treat these fields as regeneratable signals that can be coupled to localization, containment, and cleanup tasks through probes and task-specific metrics.}
 \label{fig:ocpnet_panels}
\end{figure}

\revise{Actions specify desired relative velocity, clipped as $a_t=\mathrm{clip}(\tilde a_t;v_{\max})$. With current gain $g$,}
\begin{equation}
\revise{c_t = g\,[u_{\mathrm{data}}(x_t,z_t),\,0,\,v_{\mathrm{data}}(x_t,z_t)] + c_{\mathrm{tide}}(t),}
\end{equation}
\revise{where $u_{\mathrm{data}}$ and $v_{\mathrm{data}}$ are sampled horizontal currents and $c_{\mathrm{tide}}$ is an optional synthetic robustness signal. With body velocity $\nu=[\nu_\ell^\top,\nu_a^\top]^\top$, rotation $R$, and diagonal $M,D,K_\ell,K_a,K_d$, the 6DoF engineering update is}
\begin{align}
M\dot\nu &= \tau-D\nu, &
\tau_\ell &= K_\ell(\nu_\ell^{\rm cmd}-\nu_\ell), \nonumber\\
\tau_a &= K_a(\nu_a^{\rm cmd}-\nu_a)-K_d\nu_a, &
\dot p &= R\nu_\ell+c_t .
\end{align}
\revise{Commands are transformed to the body frame; yaw follows horizontal motion while roll and pitch are stabilized. Kinematic mode directly integrates $a_t+c_t$, and 3DoF suppresses roll and pitch. The diagonal model is not vehicle-identified hydrodynamics: it omits added mass, Coriolis and restoring terms, thruster allocation, and two-way fluid coupling. Hard terrain feasibility is}
\begin{equation}
\mathcal C=\{(x,y,z)\mid m_{\mathrm{land}}(x,z)<\tau_{\mathrm{land}},\ y+\delta_{\mathrm{clr}}\le-h(x,z)\},
\end{equation}
\revise{where $m_{\mathrm{land}}$ is mask validity, $h$ is bathymetric elevation, and $\delta_{\mathrm{clr}}$ is required clearance. Masked, out-of-bounds, or clearance-violating proposals are rejected through the shared constraint path. We report SR, timeout-aware episode time $T$, $E=\sum_t\|a_t\|_2^2\Delta t$, near-collision ratio, and task-specific metrics.}

\revise{BC maps normalized goal, depth, probe, current, and task features $\phi_t$ to $\hat a_t=f_\theta(\phi_t)$ and minimizes the heuristic-action loss}
\begin{equation}
\mathcal L_{\mathrm{BC}}=N_{\mathrm{BC}}^{-1}\sum_t\|a_t^{\mathrm{heur}}-\hat a_t\|_2^2.
\end{equation}
\revise{Every $K$ steps, an LLM instead proposes a discrete assignment $g_k=\pi_{\mathrm{LLM}}(s_k)$; schema-invalid output invokes a deterministic allocator, and a shared low-level controller executes $a_t=\pi_{\mathrm{low}}(x_t,g_k)$. Appendix Sec.~\ref{app:models_ros} gives dynamics defaults, pollution semantics, and ROS mapping; Appendix Sec.~\ref{app:protocol} defines all metrics and manifests.}

\paragraph{Optional pollution fields (OCPNet).}
\label{sec:ocpnet}

\revise{Continuous containment applies a local concentration sink, whereas canonical SPC completes discrete sources after an assignment-and-dwell condition; its cleanup rate is therefore source completion, not removed fluid mass. Figure~\ref{fig:ocpnet_panels} shows the regeneratable multi-species signal.}

\section{Experiments}
\label{sec:experiments}

\revise{We evaluate OneOcean along four axes: method separation on the unified 10-task suite under 6DoF settings; difficulty and control-versus-planning gaps across the tiered task ladder; trade-offs in success, efficiency, and compute among heuristic control, BC, and local LLM planners; and transfer of the same task semantics to the external underwater-scene backend grounded by the same ocean data product.}

\subsection{Experimental setup}

\revise{The main core results use 6DoF with hard mask and clearance checks and a fixed NetCDF-derived current slice; the external backend advances a temporal export in four packaged scenes. We report SR, timeout-aware $T$, action-energy proxy $E$, near-collision ratio, and the task metrics named in the tables. Headline budgets are stated with their result tables, while separate diagnostics audit planner decisions, dynamics sensitivity, and pollution observations. All runs store their integer seed, task and dynamics configuration, dataset indices, and software and Git state. Local LLM inference assigns one 48\,GB accelerator to each process. Appendix Sec.~\ref{app:protocol} specifies the supplementary diagnostic budgets and complete manifest contract.}

\revise{Here $T$ uses success time on successful episodes and timeout on failures; $E$ is computed only from clipped vehicle commands and is not inference energy. Coll(\%) counts steps in which the closest agent pair is within the configured near-collision radius. Coverage, inspection, source completion, station error, Cross-Track Error (CTE), and reached waypoints expose progress that binary SR would hide. Tiers 1--3 jointly change disturbance, tolerance, spatial extent, and coordination settings rather than varying one scalar.}

\subsection{Core benchmark backend}

\revise{
\begin{table}[!t]
\centering
\scriptsize
\setlength{\tabcolsep}{3.0pt}
\renewcommand{\arraystretch}{1.10}
\begin{tabularx}{\columnwidth}{@{}>{\raggedright\arraybackslash}p{0.21\columnwidth}c*{4}{Y}@{}}
\toprule
\textbf{Task} & \textbf{$N$} & \textbf{Heur T2} & \textbf{BC T2} & \textbf{Heur T3} & \textbf{BC T3} \\
\midrule
\rowcolor{scenarioGray}\multicolumn{6}{c}{\textit{Navigation and Tracking}} \\
\makecell[l]{G2G} & 1 & \best{94.0} & 44.0 & \best{74.0} & 34.0 \\
\makecell[l]{SK} & 1 & \best{100.0} & 0.0 & \best{14.0} & 0.0 \\
\makecell[l]{RF} & 1 & \best{84.0} & 24.0 & \best{54.0} & 0.0 \\
\makecell[l]{DPT} & 1 & \best{84.0} & 64.0 & \best{50.0} & 0.0 \\
\rowcolor{scenarioGray}\multicolumn{6}{c}{\textit{Coordination}} \\
\makecell[l]{FT} & 10 & \best{100.0} & 40.0 & \best{20.0} & 0.0 \\
\makecell[l]{FH} & 8 & 94.0 & \best{100.0} & 4.0 & \best{4.0} \\
\rowcolor{scenarioGray}\multicolumn{6}{c}{\textit{Sensing and Pollution}} \\
\rowcolor{oursbg}
\makecell[l]{PI} & 1 & \best{100.0} & 90.0 & \best{30.0} & 14.0 \\
\rowcolor{oursbg}
\makecell[l]{AS} & 1 & \best{90.0} & 84.0 & \best{14.0} & 4.0 \\
\rowcolor{oursbg}
\makecell[l]{SPC} & 10 & \best{84.0} & 50.0 & \best{10.0} & 0.0 \\
\rowcolor{oursbg}
\makecell[l]{UPL} & 5 & 16.0 & \best{90.0} & \best{20.0} & 4.0 \\
\bottomrule
\end{tabularx}
\caption{Per-task success rates at Tier 2 and Tier 3 for the canonical 10-task suite, using 50 episodes per task and tier. Highlighted rows are revisited in the planning and external-backend transfer studies.}
\label{tab:per_task_sr_t2_t3}
\end{table}
}

\revise{
\begin{figure}[!t]
 \centering
 \subfloat[SPC: Surface Pollution Cleanup (multi-agent)]{\includegraphics[width=0.48\linewidth]{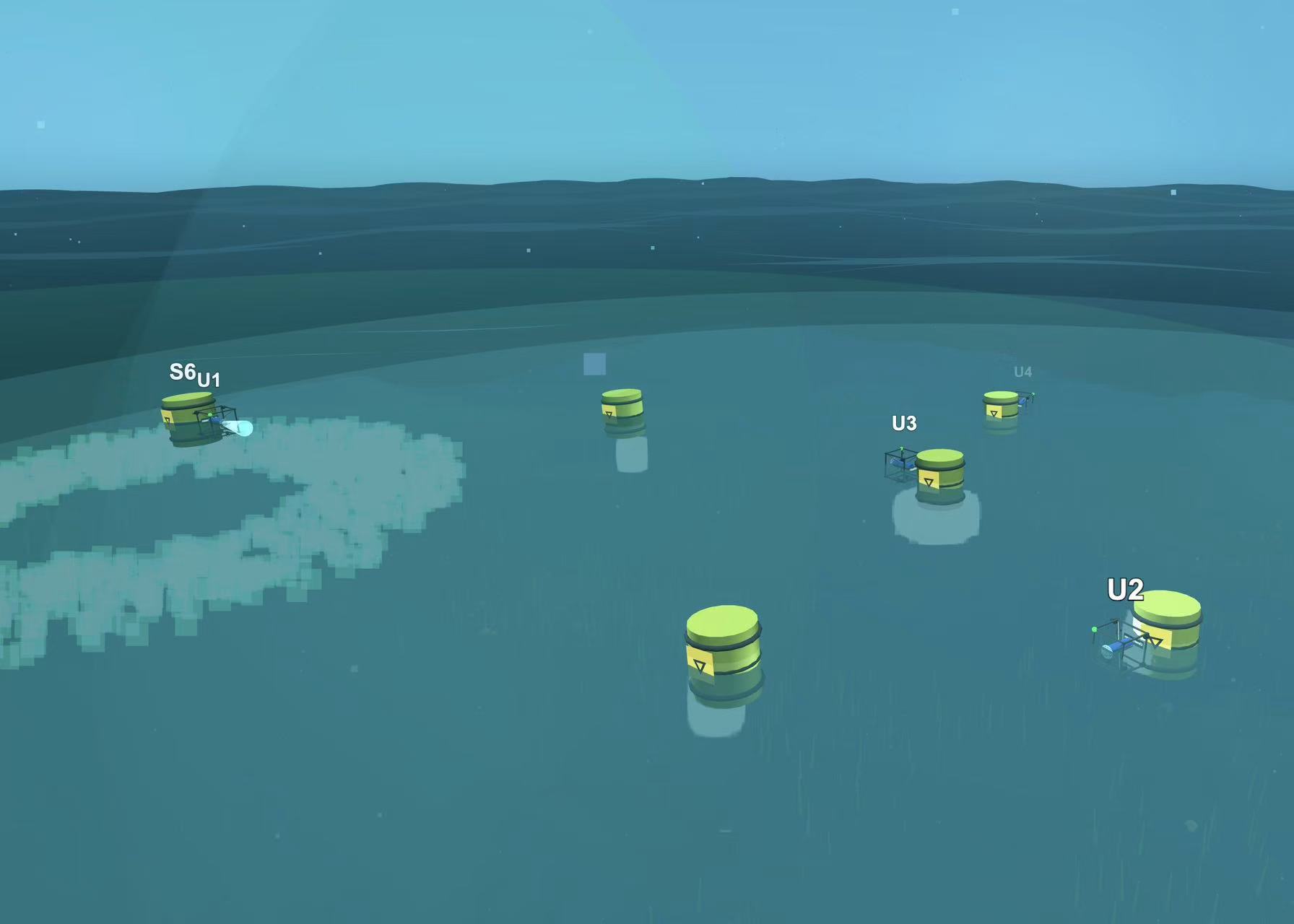}}
 \hfill
 \subfloat[UPL: Underwater Pollution Lift to Surface (5 UUVs)]{\includegraphics[width=0.48\linewidth]{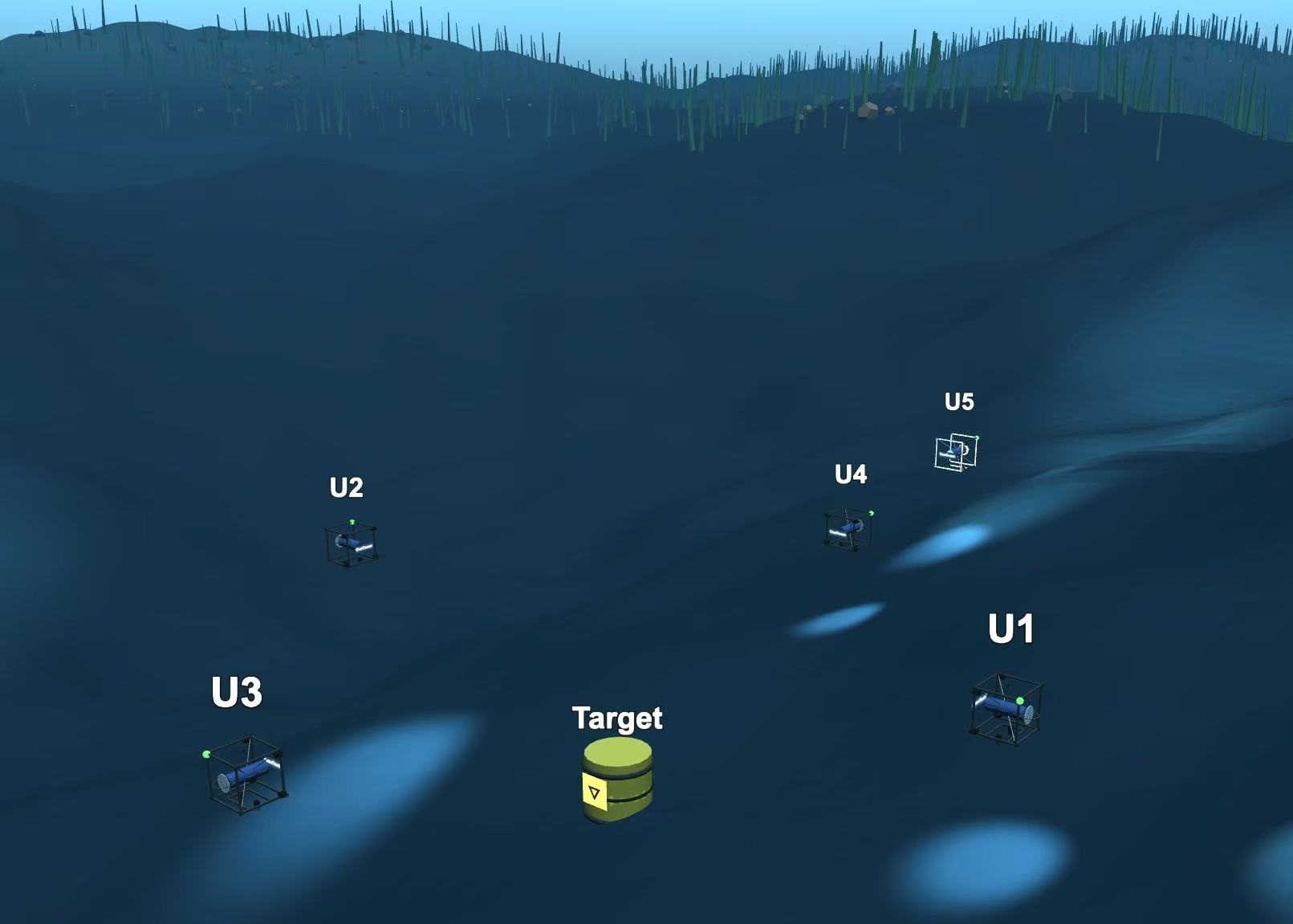}}\\
 \subfloat[AS: Area Scan over terrain]{\includegraphics[width=0.48\linewidth]{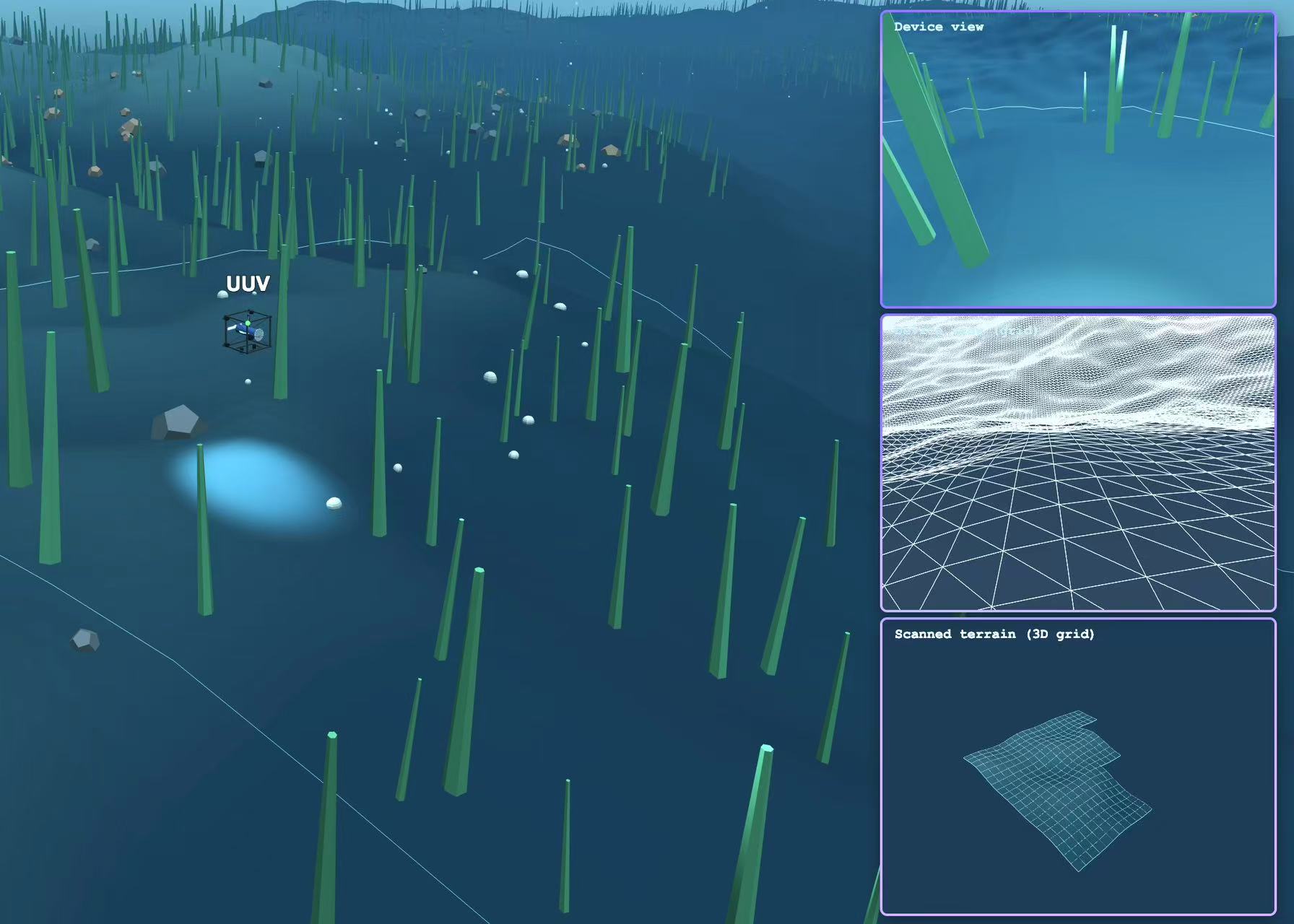}}
 \hfill
 \subfloat[FH: Fish Protection and Herding (8 UUVs)]{\includegraphics[width=0.48\linewidth]{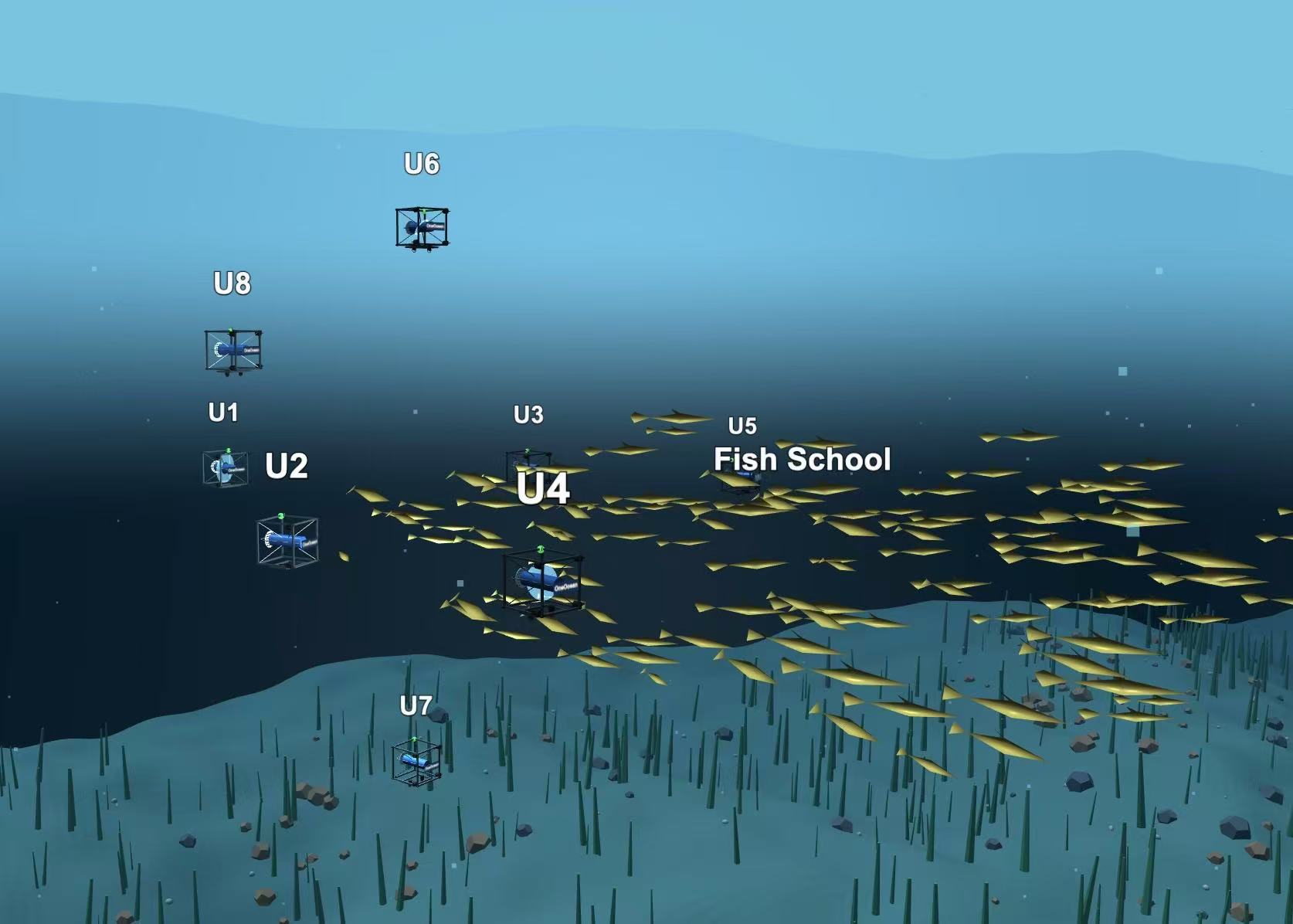}}
 \caption{Representative core-benchmark tasks used in this work's experiments. These snapshots connect the quantitative tables to the task geometry and multi-agent interactions used in the benchmark.}
 \label{fig:core_ui_off_panel}
\end{figure}
}

\revise{
\begin{table}[t]
\centering
\scriptsize
\setlength{\tabcolsep}{1.2pt}
\renewcommand{\arraystretch}{1.10}
\begin{tabularx}{\columnwidth}{@{}>{\raggedright\arraybackslash}p{0.26\columnwidth}*{7}{Y}@{}}
\toprule
\textbf{Method} & \makecell{\textbf{Avg.}\\\textbf{SR(\%)}} & \makecell{\textbf{Calls}} & \makecell{\textbf{Lat.}\\\textbf{(s)}} & \makecell{\textbf{ms per}\\\textbf{call}} & \makecell{\textbf{Prompt}\\\textbf{tok.}} & \makecell{\textbf{Output}\\\textbf{tok.}} & \makecell{\textbf{Total}\\\textbf{tok.}} \\
\midrule
\rowcolor{oursbg}
LLM-Qwen2.5-7B & \best{65.0} & 6.48 & 14.21 & 2222.1 & 3932 & 4084 & 8016 \\
LLM-Qwen2-7B & 53.3 & \best{4.71} & 11.24 & 2440.3 & 2877 & 2996 & 5873 \\
LLM-Mistral-7B & 53.3 & 6.05 & 31.38 & 4942.5 & 3762 & 4128 & 7890 \\
LLM-ChatGLM3-6B & 53.3 & 5.38 & \best{7.91} & \best{1393.6} & 3414 & \best{411} & \best{3825} \\
\rowcolor{oursbg}
Heuristic & 55.0 & \best{0.00} & \best{0.00} & \best{0.0} & \best{0} & \best{0} & \best{0} \\
LLM-Llama3-8B & 45.0 & 6.71 & 13.94 & 2117.5 & 3072 & 3218 & 6292 \\
LLM-Llama2-7B & 33.3 & 6.59 & 49.42 & 7401.8 & 4219 & 4851 & 9071 \\
BC & 21.7 & \best{0.00} & \best{0.00} & \best{0.0} & \best{0} & \best{0} & \best{0} \\
\bottomrule
\end{tabularx}
\caption{Core-benchmark planning-suite cost metrics at Tier 2 with doubled currents: uncached local-inference calls, latency, and token usage, reported alongside average success over 20 episodes per task.}
\label{tab:planning_suite_cost_t2}
\end{table}
}

\revise{Figure~\ref{fig:core_ui_off_panel} shows four representative tasks from the canonical suite. The quantitative evidence in Tables~\ref{tab:headline_macro_keymetrics} and \ref{tab:per_task_sr_t2_t3} establishes two main properties of the benchmark. First, Tier 2 is challenging but not saturated for all methods, making it useful for regression testing and planner comparison. Second, Tier 3 exposes large method gaps on both single-agent and multi-agent tasks, especially in station keeping, route following, surface cleanup, and leak inspection.}

\revise{Figure~\ref{fig:difficulty_ladder_dualpanel} makes the ladder effect explicit on a representative task subset: the left panel shows the tier-wise success profile, while the right panel isolates where the largest drops occur as the benchmark moves from Tier~1 to Tier~3.}

\revise{
\begin{figure}[!t]
 \centering
 \includegraphics[width=0.98\linewidth]{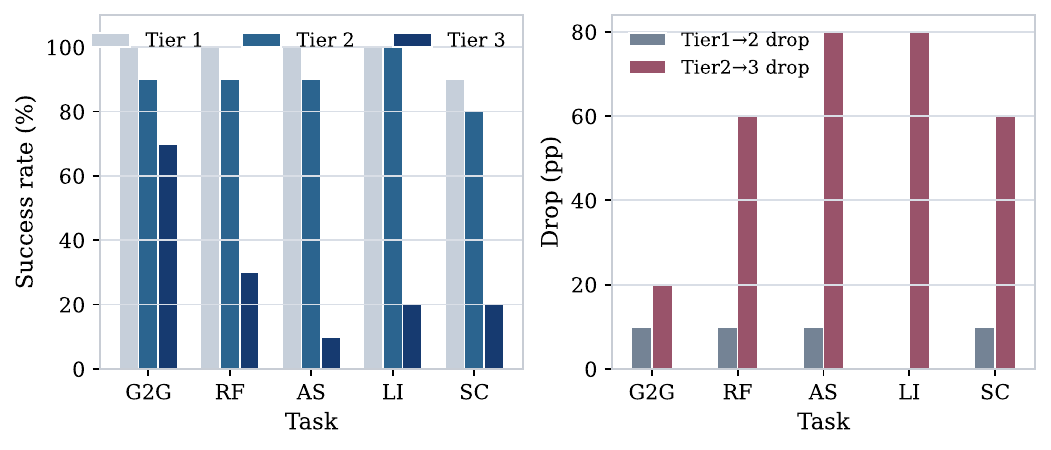}
 \caption{Difficulty-ladder summary on a representative five-task subset. Left: success rate at Tiers~1--3. Right: the corresponding success-rate drops from Tier~1$\rightarrow$2 and Tier~2$\rightarrow$3, highlighting that the hardest setting remains strongly discriminative for route following, area scan, leak inspection, and surface cleanup.}
 \label{fig:difficulty_ladder_dualpanel}
\end{figure}
}

\revise{Two observations stand out. Heuristic control remains the strongest general-purpose baseline in the core benchmark backend, especially once Tier 3 activates stricter tolerances and stronger drift. BC remains competitive on \emph{Underwater-Lift} and \emph{Fish-Herding} at Tier 2. However, it degrades sharply on tasks that require persistent disturbance rejection or coordinated geometry maintenance. Figure~\ref{fig:cleanup_scaling_dualpanel} further shows that the surface-cleanup task captures a meaningful team-size trade-off: larger teams improve success and source-completion rate, but collision risk increases and the energy budget does not scale monotonically.}

\paragraph{Planning-sensitive tasks.}

\revise{We next focus on three planning-sensitive tasks: surface cleanup, area scan, and leak inspection. We compare deterministic heuristic control, BC, and several local LLM planners. The LLM never replaces low-level control; it proposes high-level assignments or waypoint choices at a fixed stride, while the low-level controller and safety checks remain deterministic.}

\revise{Tables~\ref{tab:planning_suite_perf_t2} and \ref{tab:planning_suite_cost_t2} show a useful separation. The heuristic remains the most reliable overall planner, but several local LLMs are competitive on specific tasks: ChatGLM3-6B matches the heuristic on cleanup rate and scan coverage with lower latency than the other LLMs, while Qwen2.5-7B gives the best average success among the LLM family in this sweep.}

\revise{An instrumented three-seed audit explains why valid language output does not imply task success: fallback can invoke a competent allocator, while valid plans can still repeat covered regions or miss leaks. Thus $E$ measures executed trajectories, whereas latency and token counts measure inference cost. The LLM rows are high-level planning diagnostics rather than architecture-matched low-level controller rankings. Appendix Sec.~\ref{app:core_diagnostics} reports task-level and model-level validity, fallback, completion, token, and latency values.}

\revise{
\begin{figure}[!b]
 \centering
 \includegraphics[width=0.98\linewidth]{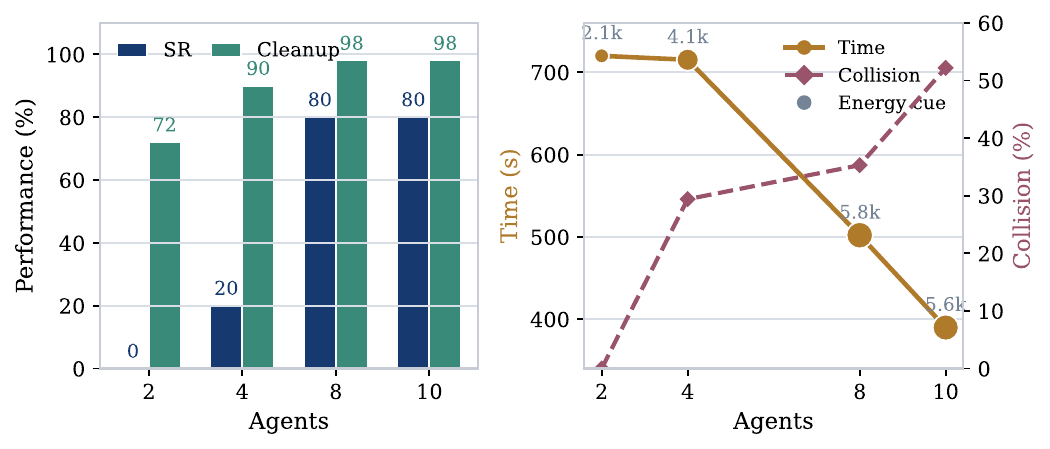}
 \caption{Surface-cleanup scaling at Tier~2. Left: success rate and cleanup rate as the number of vehicles increases. Right: completion time and near-collision ratio, with per-point annotations showing the energy proxy in thousands. The figure highlights that scaling improves cleanup outcomes, but also shifts the coordination cost profile.}
 \label{fig:cleanup_scaling_dualpanel}
\end{figure}
}

\subsection{External underwater-scene backend}

\revise{
\begin{figure}[ht]
 \centering
 \subfloat[Surface pollution containment and cleanup in a packaged ocean scene.]{\includegraphics[width=0.48\linewidth]{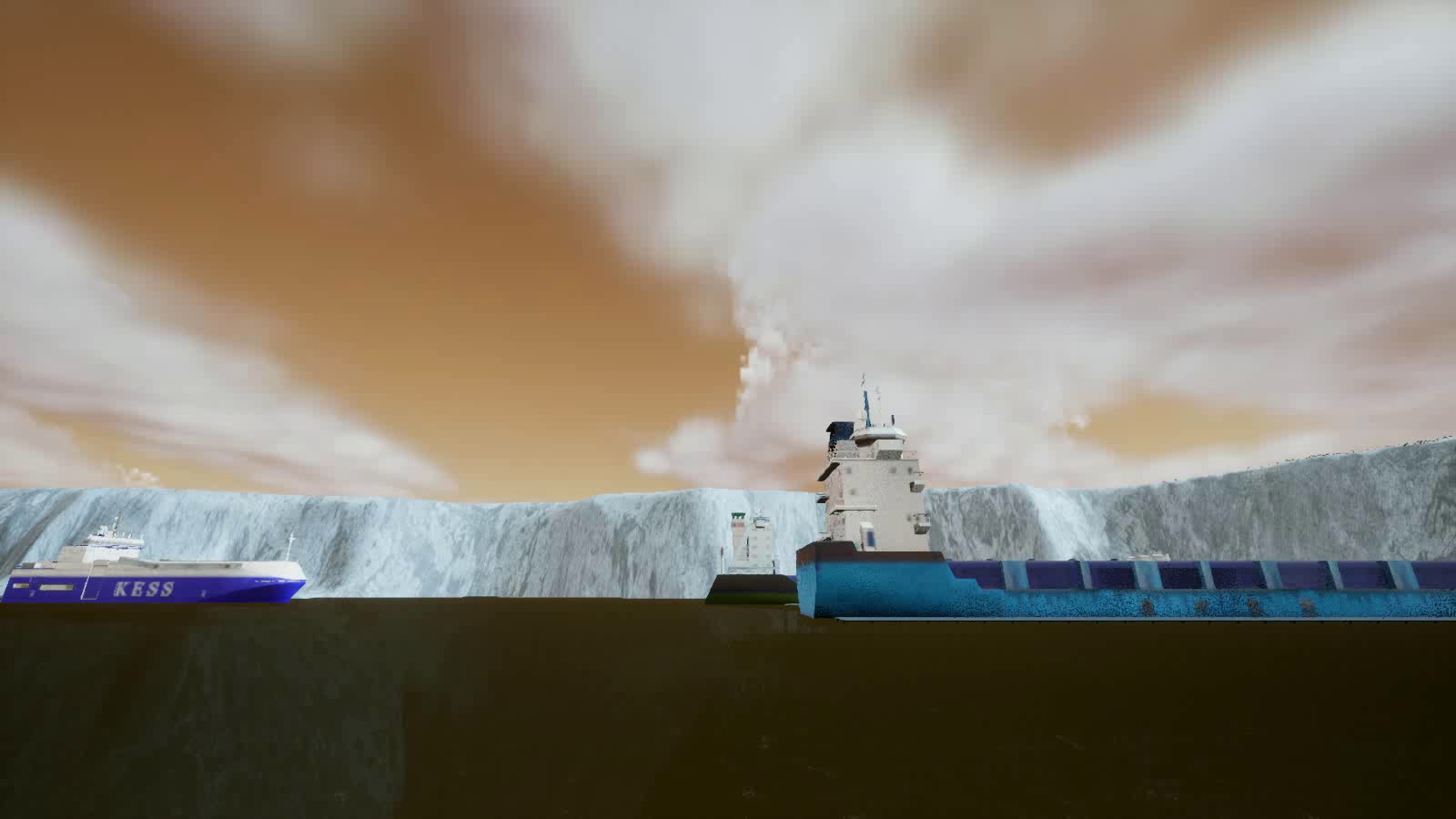}}
 \hfill
 \subfloat[Multi-agent scene-backed deployment.]{\includegraphics[width=0.48\linewidth]{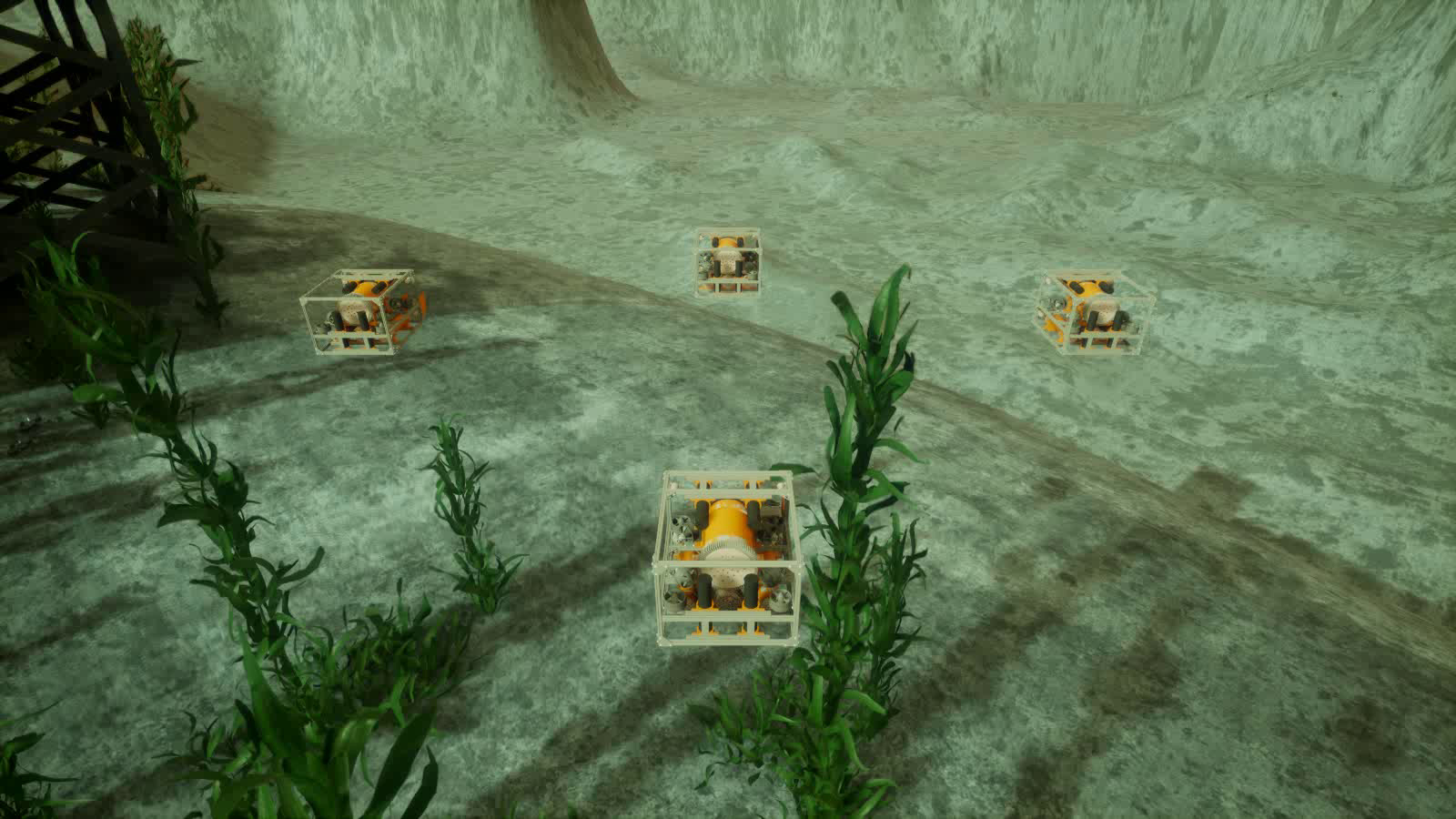}}
 \caption{Qualitative evidence from the external underwater-scene backend, which reuses packaged underwater worlds while injecting currents exported from the same combined dataset used by the core benchmark backend.}
 \label{fig:extscene_keyframes}
\end{figure}
}

\revise{Figure~\ref{fig:extscene_keyframes} and Table~\ref{tab:h3_route_generalization} show that the shared task contract remains executable but route following degrades differently across scenes. A separate diagnostic found lower localization error with OCPNet-3D under both tested current settings without reducing aggregate success. Appendix Sec.~\ref{app:external_diagnostics} provides the values, episode budget, and interpretation limits.}

\revise{
\begin{table}[t]
\centering
\scriptsize
\setlength{\tabcolsep}{1.2pt}
\renewcommand{\arraystretch}{1.12}
\begin{tabularx}{\columnwidth}{@{}>{\raggedright\arraybackslash}p{0.25\columnwidth}*{6}{Y}@{}}
\toprule
\multirow{2}{*}{\textbf{Scene}} & \multicolumn{3}{c}{\textbf{Tier 2}} & \multicolumn{3}{c}{\textbf{Tier 3}} \\
\cmidrule(lr){2-4}\cmidrule(lr){5-7}
 & \makecell{SR\\(\%)} & \makecell{WP} & \makecell{CTE\\(m)} & \makecell{SR\\(\%)} & \makecell{WP} & \makecell{CTE\\(m)} \\
\midrule
Dam & 0.0 & 1.0 & 3.06 & 0.0 & 0.0 & \best{0.87} \\
OpenWater & 20.0 & 3.1 & 2.50 & 0.0 & 0.0 & 1.91 \\
PierHarbor & 56.7 & 3.5 & \best{2.22} & 0.0 & 2.0 & 1.41 \\
SimpleUnderw. & \best{100.0} & \best{4.0} & 3.04 & \best{96.7} & \best{3.0} & 2.16 \\
\rowcolor{scenarioGray}
Mean over scenes & 44.2 & 2.92 & 2.70 & 24.2 & 1.25 & 1.59 \\
\bottomrule
\end{tabularx}
\caption{Route-following across four underwater scenes after the success-criterion fix. Each scene uses 30 episodes per difficulty. ``Waypoints'' is the mean number of reached waypoints and CTE is Cross-Track Error.}
\label{tab:h3_route_generalization}
\end{table}
}

\revise{The external underwater-scene results should be read as a portability study rather than as a replacement for core-benchmark throughput evaluation. This backend reuses realistic scenes and produces paper and web media, but it also introduces stricter scene geometry and a different disturbance hook. Both backends retain the same task semantics, data provenance, and reporting contract.}

\section{Discussion and Limitations}
\label{sec:discussion}

\revise{OneOcean's data-grounded disturbances provide repeatable stress tests across task families. Tier 2 supports regression testing, whereas Tier 3 exposes failures in tracking and coordination. No method dominates every criterion: heuristic control is strongest overall, BC remains competitive on selected structured tasks, and local LLM planners trade improved planning behavior for additional latency and token cost.}

\revise{Success alone can obscure substantial partial progress and differences in energy, collision exposure, or completion time. We therefore pair success with task-specific progress. Planner validity is reported separately from closed-loop performance because a syntactically valid assignment may be dynamically poor, while deterministic fallback can conceal a failed planner call. Backend portability is assessed through shared task contracts, data provenance, output semantics, and common constraint checks rather than direct visual-fidelity or throughput rankings between unlike simulators. A controller, vehicle, or backend is compatible only when it consumes the environment query contract, declares termination conditions and metrics, records its configuration and dataset indices, and preserves the shared constraint path.}

\revise{The validator and ROS~2 smoke checks establish interface compliance rather than physical fidelity or task performance. OneOcean remains a controlled engineering abstraction rather than a full hydrodynamic simulator. Its diagonal 6DoF update omits added mass, Coriolis and restoring forces, thruster allocation, and two-way fluid coupling. Dataset currents use a recorded relative-transport convention; the external backend applies them through a one-way disturbance hook. The \texttt{tiny} product has one near-surface layer, whereas the larger products retain 26 measured horizontal-current levels. OCPNet nevertheless repeats horizontal current over model depth and assumes zero vertical current. Appendix Sec.~\ref{app:protocol} details the remaining pollution, tide, ROS~2, and extension boundaries.}

\section{Conclusion}
\revise{OneOcean combines a unified spatiotemporal ocean-environment product, a canonical 10-task ladder, and standardized evaluation for marine robotics. Its headless and underwater-scene backends support reproducible experiments under data-grounded disturbances. Comparisons of heuristic control, BC, and local LLM planners demonstrate the suite's utility for quantitative evaluation, robustness and planner-cost analysis, and scene-backed portability studies. Future work will improve tide coupling and three-dimensional currents, broaden tasks and robot morphologies, and extend the release of derived datasets and benchmark artifacts.}

\clearpage
\bibliographystyle{aaai2027}
\bibliography{references}
\clearpage
\appendix
\setcounter{secnumdepth}{1}
\setcounter{figure}{0}
\setcounter{table}{0}
\setcounter{equation}{0}
\renewcommand{\thefigure}{A\arabic{figure}}
\renewcommand{\thetable}{A\arabic{table}}
\renewcommand{\theequation}{A\arabic{equation}}
\renewcommand{\theHfigure}{appendix.\arabic{figure}}
\renewcommand{\theHtable}{appendix.\arabic{table}}
\renewcommand{\theHequation}{appendix.\arabic{equation}}
\section*{Appendix}
\noindent This appendix details the environment representation, task objectives, motion and pollution models, and evaluation protocol. It also reports complementary studies of dynamics, disturbance strength, planner decisions, and scene-backed multi-agent behavior.
\section{Environment Product Contract and Validation}
\label{app:data}

\subsection{Sources, harmonization, and schema}

OneOcean combines bathymetry from the General Bathymetric Chart of the Oceans (GEBCO) \cite{GEBCO2025} with horizontal currents from the Copernicus Marine Service Global Ocean Physics Analysis and Forecast (GOPAF) product \cite{copernicus,CopernicusGOPAF2025}. The resulting Network Common Data Form (NetCDF) product is harmonized by cropping the requested region, retaining returned time and depth coordinates, normalizing coordinate names and units, interpolating variables onto one latitude--longitude grid, preserving invalid terrain and NoData cells, and recording the transformations. Harmonization does not mean inventing unavailable depth layers.

\begin{figure*}[t]
\centering
\includegraphics[width=0.98\textwidth]{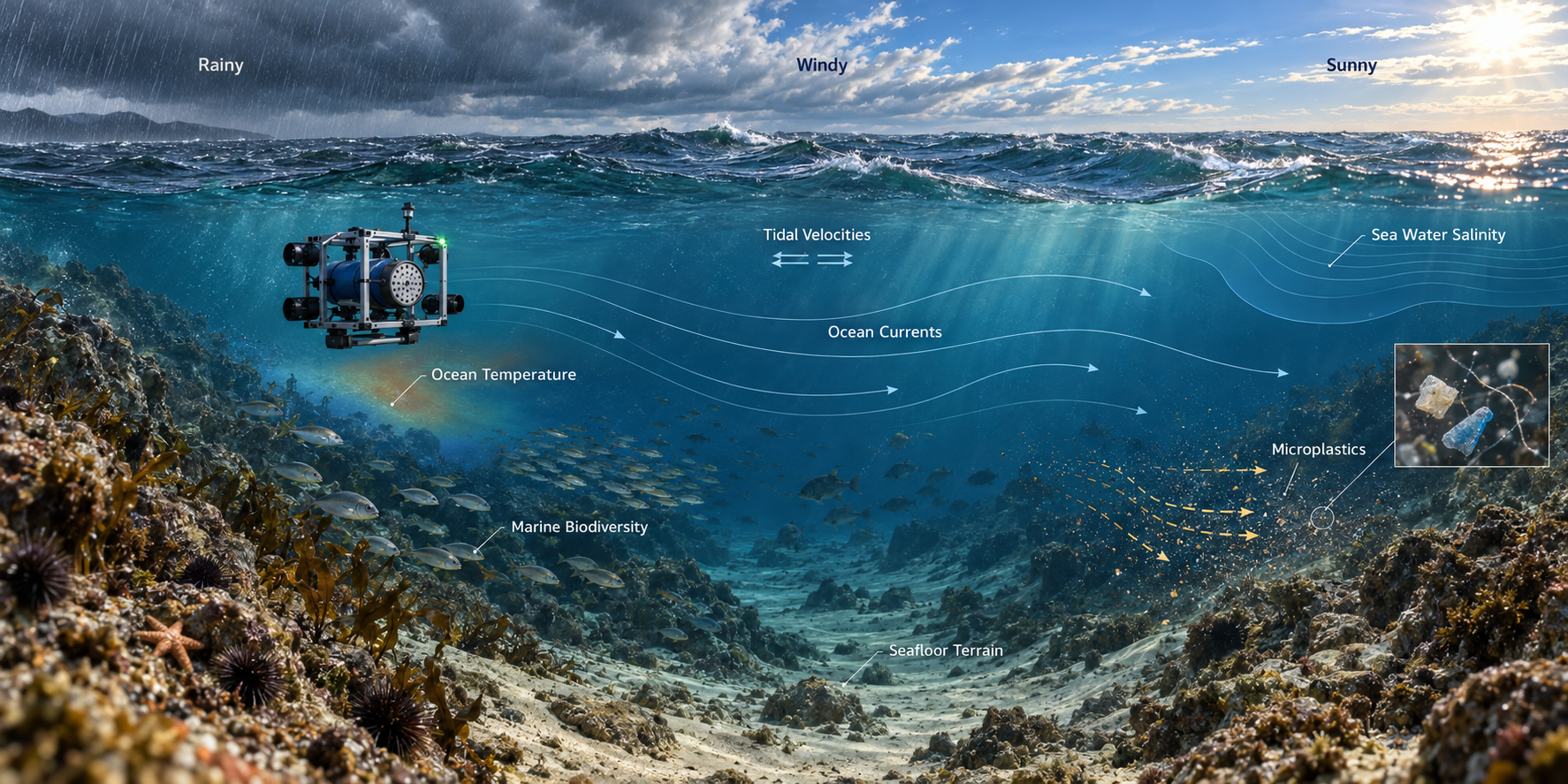}
\caption{Conceptual view of the environmental channels and ecological signals represented by, or targeted for coupling with, OneOcean. It illustrates terrain, currents, tidal velocity, thermohaline variables, weather, biodiversity, and microplastics; it is not a scale diagram and does not imply that every release variant contains every depicted field.}
\label{fig:app_environment_overview}
\end{figure*}

\begin{table*}[t]
\centering
\scriptsize
\setlength{\tabcolsep}{4pt}
\renewcommand{\arraystretch}{1.10}
\begin{tabularx}{\textwidth}{l l l X}
\toprule
\textbf{Canonical field} & \textbf{Dimensions} & \textbf{Unit or Encoding} & \textbf{Contract} \\
\midrule
Latitude & $H$ & degrees north & Strictly monotonic geographic coordinate. \\
Longitude & $W$ & degrees east & Strictly monotonic geographic coordinate. \\
Time & $T$ & Climate and Forecast time unit & Monotonic timestamps retained from the source product. \\
Depth & $D$ & m, positive down & Actual source levels; a one-level product is valid but not a resolved 3D flow. \\
Eastward and northward current & $T\!\times\!D\!\times\!H\!\times\!W$ & m\,s$^{-1}$ & Base pair $(u,v)$ or tide-combined pair $(u_{\mathrm{total}},v_{\mathrm{total}})$. \\
Elevation & $H\!\times\!W$ & m relative to sea surface & Bathymetric elevation aligned to the current grid. \\
Validity mask & $H\!\times\!W$ & binary & Invalid terrain, crop-exterior, or NoData cells; not necessarily a coastline classifier. \\
\bottomrule
\end{tabularx}
\caption{Canonical environment-product schema. The validator accepts documented aliases but maps every accepted product to these semantics.}
\label{tab:app_data_contract}
\end{table*}

The storage merge fills masked elevation cells with zero, so elevation must always be interpreted together with the validity mask. A zero at an invalid cell is not a physical seabed measurement. In the core simulation frame, $(x,y,z)=$(east, depth-down, north), and the dataset vector $(u,v)$ maps to $(u,0,v)$. Longitude and latitude are converted to local metric $x$ and $z$ about the recorded lower-left geographic origin.

\subsection{Depth and tide treatment}

The realized \texttt{tiny} request contains one level at 0.494\,m. The \texttt{scene} and \texttt{public} requests cover 0--200\,m and retain 26 source levels from 0.494 to 186.126\,m. The pipeline records returned coordinates rather than relabeling a nearest source level as the requested value. Consequently, \texttt{tiny} is a time-varying 2D or 2.5D current product, whereas the other two products provide measured horizontal current at multiple discrete depths.

Only the current \texttt{scene} release includes tide components. Hourly tide samples are aligned to the base product with nearest-time selection and attenuated as $\exp(-d/z_0)$ with $z_0=50$\,m before forming the total current pair. This is a documented engineering approximation chosen for data volume and simulation convenience. It discards high-frequency structure on coarser time grids and is neither a validated 3D tidal model nor suitable for studies that require phase-accurate tidal dynamics.

\subsection{Validator evidence}

The release validator checks required fields, aliases, dimensions, monotonic coordinates, units, finite sampled values, mask semantics, provenance, and optional round-trip fidelity between NetCDF and a NumPy Archive (NPZ). Table~\ref{tab:app_validator} summarizes the July 21, 2026 audit of the already-generated products.

\begin{table}[t]
\centering
\scriptsize
\setlength{\tabcolsep}{3pt}
\renewcommand{\arraystretch}{1.10}
\begin{tabularx}{\columnwidth}{l X c c l}
\toprule
\textbf{Variant} & \textbf{Current pair} & \textbf{Errors} & \textbf{Warn.} & \textbf{Round-trip} \\
\midrule
\texttt{tiny} & \texttt{uo}, \texttt{vo} & 0 & 3 & Exact \\
\texttt{scene} & \texttt{utotal}, \texttt{vtotal} & 0 & 4 & Not run \\
\texttt{public} & \texttt{uo}, \texttt{vo} & 0 & 2 & Not run \\
\bottomrule
\end{tabularx}
\caption{Environment-product validation. All variants are \texttt{PASS\_WITH\_WARNINGS}; warnings concern legacy metadata rather than numeric or schema errors. Round-trip denotes the fixed \texttt{tiny} $(t=0,d=0)$ NetCDF-to-NPZ check.}
\label{tab:app_validator}
\end{table}

The legacy warnings identify missing depth or elevation units, total-current units, or mask flag meanings depending on the variant. Regeneration with the updated pipeline writes these attributes. The audited \texttt{tiny} slice preserved coordinates and currents at absolute tolerance $10^{-6}$.

\subsection{Portable provenance}

Every regenerated NetCDF records source product identifiers, requested and actual region, time, and depth, tide inclusion and depth profile, interpolation and missing-value handling, and generation software. Companion variant manifests use paths relative to the variant directory; private workspace paths are excluded. This contract allows a custom product to fail early before an episode is launched and lets every run retain the selected time and depth indices.

\subsection{Geographic sampling within an episode}

Let $(\phi_0,\lambda_0)$ be an episode's geographic origin and $p=(x,y,z)$ its local position in meters. The core sampler uses
\begin{align}
\phi(p)&=\phi_0+z/L,\\
\lambda(p)&=\lambda_0+x/[L\max(0.1,\cos\phi_0)],
\end{align}
where $L=111132$\,m per degree and the cosine evaluates latitude in radians. At reset, the origin is sampled within the available geographic extent to place the simulation tile on the environmental grid. The default tile spans $600\times600$\,m horizontally, with vehicle depth bounded between 2 and 18\,m. Environmental coordinates retain the source grid spacing.

For grid coordinates $\{\phi_j\}$ and $\{\lambda_k\}$, the sampled indices are
\begin{equation}
j^*=\arg\min_j|\phi_j-\phi(p)|,\qquad
k^*=\arg\min_k|\lambda_k-\lambda(p)|.
\end{equation}
The current at $p$ is $g(u_{j^*k^*},0,v_{j^*k^*})$, where $g$ is the current gain. Bathymetry and validity use the same indices. Missing current samples are set to zero, while terrain feasibility is determined separately by the validity mask and seabed clearance. Core episodes query a fixed time and depth slice.

\subsection{Temporal forcing in the scene backend}

The scene backend uses an exported current sequence at a selected geographic location and the nearest available source depth. Let $T_c$ be its number of temporal samples, $s$ the episode seed, and $\alpha$ the configured sequence-advance rate. At simulation time $t$, the sampled index is
\begin{equation}
i(t)=\big[(s\bmod T_c)+\lfloor\alpha t\rfloor\big]\bmod T_c.
\end{equation}
Setting $\alpha=0$ holds the initial sample; a positive value advances the sequence and wraps at its end. For a daily export, $\alpha$ gives source days per simulated second. The sampled vector, multiplied by the current gain, drives the scene disturbance. Spatial slices and temporal sequences provide complementary conditions for evaluating disturbance rejection.

\section{Canonical Task Contracts}
\label{app:tasks}

Every task declares an objective, observation contract, feasibility constraints, success condition, and primary metric. Shared observations include pose and relative target cues, local current, and task-specific probes; shared feasibility checks include domain bounds, terrain validity, seafloor clearance, and configured team size.

\begin{description}
\item[Go-to-Goal Under Currents (G2G).] Reach a goal while rejecting data-grounded drift; success requires entering the goal tolerance, and final goal error is primary.
\item[Station Keeping (SK).] Hold position under disturbance; station RMS error measures sustained deviation.
\item[Route Following (RF).] Visit an ordered waypoint route; cross-track error and reached waypoints expose partial progress.
\item[Depth-Profile Tracking (DPT).] Follow a target depth profile during motion; depth-tracking error is primary.
\item[Area Scan (AS).] Cover a bounded region over 3D terrain; grid coverage fraction measures progress.
\item[Pipeline Inspection (PI).] Follow a pipeline corridor and encounter leak cues; the inspection score is the detected-leak fraction, with route progress recorded separately.
\item[Formation Transit (FT).] Move a team of 2--10 vehicles while maintaining relative geometry; formation error and collision ratio capture coordination quality.
\item[Surface Pollution Cleanup (SPC).] Assign 2--10 vehicles to discrete surface sources and satisfy per-source dwell requirements; cleanup rate is completed-source fraction.
\item[Underwater Pollution Lift (UPL).] Coordinate five vehicles around a payload and lift it to the surface; lift completion defines success.
\item[Fish Protection and Herding (FH).] Use eight vehicles to contain and escort a fish school under currents; centroid progress toward the escort goal defines success.
\end{description}

\subsection{Exact progress metrics}

Several task metrics are normalized ratios with distinct denominators. Let $\mathcal V$ be the visited scan cells in an $H\times W$ grid, $N_{\mathrm{det}}$ the detected leaks among $N_{\mathrm{leak}}$ placed leaks, and $N_{\mathrm{done}}$ the sources whose dwell requirement is complete among $N_{\mathrm{src}}$ sources. The released evaluator computes
\begin{align}
\mathrm{Cov}_{\mathrm{AS}}
  &= \frac{|\mathcal V|}{HW}, \label{eq:as_progress}\\
\mathrm{Score}_{\mathrm{PI}}
  &= \frac{N_{\mathrm{det}}}{N_{\mathrm{leak}}}, \label{eq:pi_progress}\\
\mathrm{Rate}_{\mathrm{SPC}}
  &= \frac{N_{\mathrm{done}}}{N_{\mathrm{src}}}, \label{eq:spc_progress}\\
e_{\mathrm{FT}}
  &= \frac{1}{N}\sum_{i=1}^{N}
     \left\|p_i-(p_g+r_i)\right\|_2, \label{eq:ft_error}\\
\mathrm{Prog}_{\mathrm{FH}}(t)
  &= \operatorname{clip}\!\left(1-\frac{d_t}{d_0},0,1\right). \label{eq:fh_progress}
\end{align}
Here $p_g+r_i$ is agent $i$'s assigned formation goal, while $d_t$ and $d_0$ are the current and initial fish-centroid distances to the escort goal. These quantities expose partial progress but are not interchangeable physical measurements.

Tiers 1, 2, and 3 preserve these contracts while jointly changing current gain, tolerances, route or coverage extent, sensing difficulty, and coordination tightness. They are therefore named presets rather than a claim that one scalar fully orders task difficulty. Navigation and tracking tasks primarily stress disturbance rejection, sensing and inspection tasks stress spatial coverage and event encounters, and coordination and pollution tasks add team geometry, assignment, or dwell-time coupling.

\subsection{State-dependent completion}

\paragraph{Navigation and station keeping.}
Let $d_n=\min_i\|p_i^n-p_g\|_2$. G2G completes when $d_n$ enters the goal radius. SK requires consecutive qualifying steps: its hold counter follows
\begin{equation}
h_{n+1}=\begin{cases}h_n+1,&d_n\leq r,\\0,&d_n>r.\end{cases}
\end{equation}
Success occurs when $h_{n+1}$ reaches the prescribed hold length. Consequently, an excursion outside the station radius restarts the hold interval. RF and DPT maintain an ordered waypoint index, advance it when a vehicle enters the active waypoint radius, and evaluate completion from the updated index and that step's waypoint distance. DPT additionally requires the team-mean absolute depth error relative to the evaluated waypoint to satisfy its depth tolerance. FT applies its tolerance to every assigned formation goal: $\max_i\|p_i-(p_g+r_i)\|_2\leq r$.

\paragraph{Area coverage and inspection.}
AS stores a persistent binary grid of visited cells. With sensor radius $r_s$ and cell width $\ell$, a vehicle marks a square neighborhood extending $\lfloor r_s/\ell\rfloor$ cells in each horizontal direction, clipped at the grid boundary. Cells are counted once. A lawnmower route supplies intermediate goals, while completion depends on the coverage fraction in Equation~\ref{eq:as_progress}. PI maintains a separate detection flag for each leak. A flag becomes permanent when any vehicle enters the leak's detection radius; the first encounter time is recorded. Detection of all leaks completes the task, and the waypoint index records traversal progress separately.

\paragraph{Accumulated service in surface cleanup.}
For source $j$, let $a_i^n$ denote agent $i$'s assigned source and $q$ the required number of agents. Its accumulated service time is
\begin{align}
b_j^n&=\sum_i\mathbf 1[a_i^n=j]\,
                 \mathbf 1[\|p_i^n-s_j\|_2\leq r],\\
\tau_j^{n+1}&=\tau_j^n+\Delta t\,\mathbf 1[b_j^n\geq q].
\end{align}
A source completes when $\tau_j$ reaches its dwell threshold. Service already accumulated is retained when vehicles leave; this differs from the consecutive hold in SK. Only vehicles assigned to that source contribute to its service count. The task completes once every source is finished. The hard preset couples assignment with rendezvous by requiring two agents to service a source simultaneously.

\paragraph{Attachment and collective lifting.}
UPL has four phases: approach, lift-off, fifth-agent joining, and ascent. Each of the five agents acquires a persistent attachment flag upon entering the payload attachment radius. Attachment of the first four agents starts the lift-off hold; completing that hold enables joining, and attachment of the fifth agent enables ascent. During the lifting phases the payload follows the attached team's mean position with a phase-dependent depth update. Completion requires the ascent phase, all five attachments, and payload depth at or above the designated surface threshold.

\paragraph{Fish-school escort.}
FH evolves a school under current drift and a herding response to nearby vehicles. Its progress uses the horizontal distance between the school centroid and escort goal. The eight-vehicle team succeeds when either the normalized progress in Equation~\ref{eq:fh_progress} reaches its target or the centroid enters the goal radius.

\begin{table*}[t]
\centering\scriptsize
\setlength{\tabcolsep}{4pt}
\renewcommand{\arraystretch}{1.13}
\begin{tabularx}{\textwidth}{l X X r r}
\toprule
\textbf{Task} & \textbf{Tier 2 parameters} & \textbf{Tier 3 parameters} & \multicolumn{2}{c}{\textbf{Step budget}}\\
 & & & \textbf{Tier 2} & \textbf{Tier 3}\\
\midrule
G2G & Goal radius 4\,m & Goal radius 1.3\,m & 280 & 300\\
SK & Radius 2.6\,m; hold 110 steps & Radius 1.3\,m; hold 240 steps & 360 & 560\\
RF & 12 waypoints; radius 3.2\,m & 20 waypoints; radius 1.6\,m & 460 & 560\\
DPT & 12 waypoints; radius 3.6\,m; depth tolerance 1.2\,m & 18 waypoints; radius 1.8\,m; depth tolerance 0.7\,m & 560 & 720\\
FT & Formation radius 36\,m; goal tolerance 4\,m & Formation radius 70\,m; goal tolerance 1\,m & 600 & 550\\
SPC & 12 sources; radius 3.8\,m; dwell 8\,s; one agent & 6 sources; radius 2.6\,m; dwell 6\,s; two agents & 720 & 900\\
UPL & Attach radius 4\,m; hold 3.5\,s; surface depth 2.5\,m & Attach radius 3\,m; hold 4\,s; surface depth 2.5\,m & 720 & 880\\
FH & 220 fish; progress 0.995 or goal radius 12\,m & 160 fish; progress 0.99 or goal radius 7\,m & 650 & 900\\
AS & Cell 75\,m; sensor radius 35\,m; coverage 0.82 & Cell 60\,m; sensor radius 70\,m; coverage 0.95 & 2000 & 1700\\
PI & 14 waypoints; 6 leaks; detection radius 5\,m & 20 waypoints; 8 leaks; detection radius 3.6\,m & 1500 & 1500\\
\bottomrule
\end{tabularx}
\caption{Task-specific parameters for the two principal evaluation tiers. Durations follow from the step budget and simulation timestep; the default $\Delta t$ is 1\,s. UPL uses five vehicles and FH uses eight.}
\label{tab:app_task_thresholds}
\end{table*}

\section{Implementation Defaults, Pollution, and ROS 2}
\label{app:models_ros}

\subsection{Serialized engineering defaults}

The main paper defines current coupling, action clipping, diagonal engineering dynamics, and terrain feasibility. Table~\ref{tab:app_dynamics_defaults} records the numerical defaults for the Six-Degree-of-Freedom (6DoF) mode serialized in the released run specifications so that those equations can be reproduced without relying on hidden simulator state.

\begin{table*}[t]
\centering
\scriptsize
\setlength{\tabcolsep}{4pt}
\renewcommand{\arraystretch}{1.08}
\begin{tabularx}{\textwidth}{l r X l r X}
\toprule
\textbf{Field} & \textbf{Default} & \textbf{Meaning} & \textbf{Field} & \textbf{Default} & \textbf{Meaning} \\
\midrule
$\Delta t$ & 1.0 s & Outer step & $v_{\max}$ & 1.2 m\,s$^{-1}$ & Relative speed cap \\
$M_\ell$ & $(12,12,12)$ & Linear inertia proxy & $M_a$ & $(6,6,6)$ & Angular inertia proxy \\
$D_\ell$ & $(8,8,8)$ & Linear damping & $D_a$ & $(3,3,3)$ & Angular damping \\
$K_\ell$ & $(18,18,18)$ & Linear tracking gain & $K_a$ & $(10,10,10)$ & Angular tracking gain \\
$K_d$ & $(2,2,2)$ & Angular damping gain & $\omega_{\max}$ & $(1.2,1.2,1.2)$ rad\,s$^{-1}$ & Angular-rate cap \\
$k_{\mathrm{yaw}}$ & 2.0 & Yaw-rate gain & $k_{\mathrm{att}}$ & 2.0 & Roll and pitch stabilization \\
$\theta_{\max}$ & 0.6 rad & Roll and pitch limit & & & \\
\bottomrule
\end{tabularx}
\caption{Default 6DoF engineering parameters serialized in each run specification.}
\label{tab:app_dynamics_defaults}
\end{table*}

Each run also records whether it uses direct kinematic integration, the diagonal 3DoF mode, or the bounded-attitude 6DoF mode, preventing mode selection from being inferred from output-directory names.

\subsection{Discrete motion update and terrain feasibility}

The action $a_n$ specifies desired velocity relative to the water in the world frame. Define the speed projection $\Pi_v(a)=a\min(1,v_{\max}/\|a\|_2)$ for nonzero $a$ and $\Pi_v(0)=0$. In kinematic mode the candidate position is
\begin{equation}
\widetilde p_{n+1}=p_n+\Delta t[\Pi_v(a_n)+c_n],
\end{equation}
where $c_n$ is the sampled world-frame current, including the configured disturbance terms.

For the dynamic modes, let $R_n$ map body vectors to the world frame, $\nu_n=(v_n,\omega_n)$ be body-frame linear and angular velocity, and $\omega_n^{\rm cmd}$ the desired angular rate. The latter turns yaw toward the horizontal action direction and stabilizes roll and pitch toward zero. The force and moment commands are
\begin{align}
f_n&=K_\ell[R_n^\top\Pi_v(a_n)-v_n],\\
m_n&=K_a(\omega_n^{\rm cmd}-\omega_n)-K_d\omega_n.
\end{align}
With the diagonal parameters in Table~\ref{tab:app_dynamics_defaults}, the velocity update is
\begin{align}
\widetilde v_{n+1}&=v_n+\Delta t M_\ell^{-1}(f_n-D_\ell v_n),\\
\widetilde\omega_{n+1}&=\omega_n+\Delta t M_a^{-1}(m_n-D_a\omega_n).
\end{align}
Linear speed is projected with $\Pi_v$ and angular rates are clipped componentwise before advancing position:
\begin{equation}
\widetilde p_{n+1}=p_n+\Delta t[R_n v_{n+1}+c_n].
\end{equation}
Attitude advances with the clipped angular rates; yaw is wrapped to its principal interval and roll and pitch are bounded by $\theta_{\max}$. The 3DoF mode holds roll and pitch, and their rates, at zero.  Position uses the rotation at the start of the step and the updated body velocity.

The candidate position is first clamped to the configured spatial bounds. At this candidate location, valid water requires an admissible mask cell and depth no greater than seabed depth minus the clearance margin, whose default is 1\,m. A terrain-infeasible candidate retains the previous position and increments the constraint-violation count; in dynamic modes it also resets relative linear velocity to zero. Near-collision events are measured from pairwise distances after motion. They contribute to the coordination metric without altering the motion update. Task state and completion are then evaluated on the resulting vehicle positions.

\subsection{Pollution fields and cleanup semantics}

The analytic probe field is a current-advected Gaussian,
\begin{equation}
C(p,t)=m(t)C_{\mathrm{peak}}\exp[-\|p-\mu(t)\|^2/(2\sigma^2)],
\end{equation}
with default $\sigma=35$\,m and $C_{\mathrm{peak}}=1$. An agent in the 35\,m continuous-containment neighborhood applies $m(t+\Delta t)=m(t)\max(0,1-0.12\Delta t)$.

\subsubsection{Three-dimensional transport and transformation model}

The Ocean Current Prediction Network (OCPNet) pollution engine evolves each configured species $s$ through the implemented balance
\begin{align}
\frac{\partial C_s}{\partial t}={}&
-\mathbf u\!\cdot\!\nabla C_s
+\nabla\!\cdot\!(K_s\nabla C_s)-\lambda_s C_s \nonumber\\
&+\sum_r \nu_{sr}R_r+\mathcal U_s+\mathcal P_s+S_s-Q_s,
\label{eq:pollution_balance}\\
R_r={}&k_{r0}\exp[\alpha_r(T-298.15)]f_r(\mathrm{pH}) \nonumber\\
&\times\prod_{j\in\mathcal R_r}C_j,
\label{eq:pollution_reaction}\\
f_r(\mathrm{pH})={}&
\max\!\left(1+a_r\Delta\mathrm{pH}
+b_r\Delta\mathrm{pH}^2,0\right). \nonumber
\end{align}
Here $\Delta\mathrm{pH}=\mathrm{pH}-7$, $C_s$ is concentration, $\mathbf u=(u,v,w)$ is velocity, $K_s$ is diffusivity, $\lambda_s$ is first-order decay, and $\nu_{sr}$ is a signed stoichiometric coefficient. The optional terms $\mathcal U_s$ and $\mathcal P_s$ represent biological uptake and precipitation, while $S_s$ and $Q_s$ collect externally configured sources and sinks.

The code computes spatially varying diffusivity from temperature $T$, wave velocity $U_w$, and salinity $S$ as
\begin{align}
K_T & = K_{s0}\,[1+0.02(T-273.15)], \nonumber\\
K_w & = K_T+0.1U_w, \nonumber\\
K_s & = K_w\,[1+0.05(1-S/35)].
\label{eq:pollution_diffusivity}
\end{align}
These coefficients reproduce the released numerical parameterization rather than a calibrated turbulence closure. When decay is enabled, its implemented environmental dependence is
\begin{align}
\lambda_s={}&\lambda_{s0}
\exp\!\left(-\frac{E_{a,s}}{RT}\right)
f_s(\mathrm{pH})f_s(O), \label{eq:pollution_decay}\\
f_s(\mathrm{pH})={}&
\max(1+a_s\Delta\mathrm{pH}+b_s\Delta\mathrm{pH}^2,0), \nonumber\\
f_s(O)={}&\max(1+c_sO+d_sO^2,0), \nonumber
\end{align}
where the exponential factor is omitted when $E_{a,s}=0$, $R$ is the gas constant, and $O$ is dissolved oxygen. Biological uptake, when configured, contributes the Michaelis--Menten sink
\begin{equation}
\mathcal U_s=-V_{\max,s}e^{\alpha_{T,s}(T-298.15)}
(1+\alpha_{L,s}L)\frac{C_s}{K_{1/2,s}+C_s}B,
\label{eq:pollution_biouptake}
\end{equation}
with light $L$ and phytoplankton biomass $B$. For a configured cation--anion precipitation pair, the saturation and rate terms are
\begin{align}
\mathrm{SI} & = \frac{C_M C_A}{K_{\mathrm{sp}}}, \nonumber\\
r_p & = k_p e^{\alpha_p(T-298.15)}f_p(\mathrm{pH})
\max(\mathrm{SI}-1,0),
\label{eq:pollution_precipitation}
\end{align}
which subtract $r_p$ from both dissolved species and add it to the configured precipitate before its vertical settling term is applied.

For transport, the implementation uses a first-order upwind difference. Along $x$, for example,
\begin{equation}
(\delta_x^{\mathrm{up}}C)_i=
\begin{cases}
(C_i-C_{i-1})/\Delta x, & u_i\geq0,\\
(C_{i+1}-C_i)/\Delta x, & u_i<0,
\end{cases}
\label{eq:pollution_upwind}
\end{equation}
and the centered flux form of the corresponding diffusion term is
\begin{equation}
(\mathcal D_x C)_i=
\frac{K_i(C_{i+1}-C_i)-K_{i-1}(C_i-C_{i-1})}{\Delta x^2}.
\label{eq:pollution_centered_diffusion}
\end{equation}
The same constructions are applied along $y$ and $z$, and explicit Euler combines every active process:
\begin{align}
C_s^{n+1}&=C_s^n+\Delta t\,\mathcal F_s^n,
\label{eq:pollution_euler}\\
\mathcal F_s={}&\mathcal A_s+\mathcal D_s-\lambda_sC_s
+\sum_r\nu_{sr}R_r \nonumber\\
&+\mathcal U_s+\mathcal P_s+S_s-Q_s . \nonumber
\end{align}
The solver rejects velocity fields that violate
\begin{equation}
\max\!\left(
\frac{\|u\|_\infty\Delta t}{\Delta x},
\frac{\|v\|_\infty\Delta t}{\Delta y},
\frac{\|w\|_\infty\Delta t}{\Delta z}
\right)\leq1,
\label{eq:pollution_cfl}
\end{equation}
and the diffusion module exposes the conservative helper bound $\Delta t_{\mathrm{diff}}=\tfrac12\min(\Delta x,\Delta y,\Delta z)^2/\|K_s\|_\infty$. Dirichlet, Neumann, periodic, and open boundaries are configurable. The general three-dimensional update does not clip negative concentrations, so stable timesteps and physically consistent process parameters remain the user's responsibility.

The benchmark wrapper uses a $28\times28\times10$ grid, a 2\,s internal step, base diffusivity $8\times10^{-8}$, point-source emission rate 0.02, 8\,m sink radius, and sink strength 0.15\,s$^{-1}$. Its default benchmark case activates transport, diffusion, the point source, and the robot sink; the other transformation terms above are modular capabilities and are not silently activated. Dataset horizontal current is resampled and repeated over model depth with zero vertical current. This is an explicit approximation, not a measured three-dimensional velocity profile. The robot sink multiplies cells in its local grid neighborhood by $\max(0,1-0.15\Delta t)$ at each internal update.

\subsubsection{Dataset-driven surface multispecies proxy}

\begin{figure*}[t]
\centering
\includegraphics[width=0.98\textwidth]{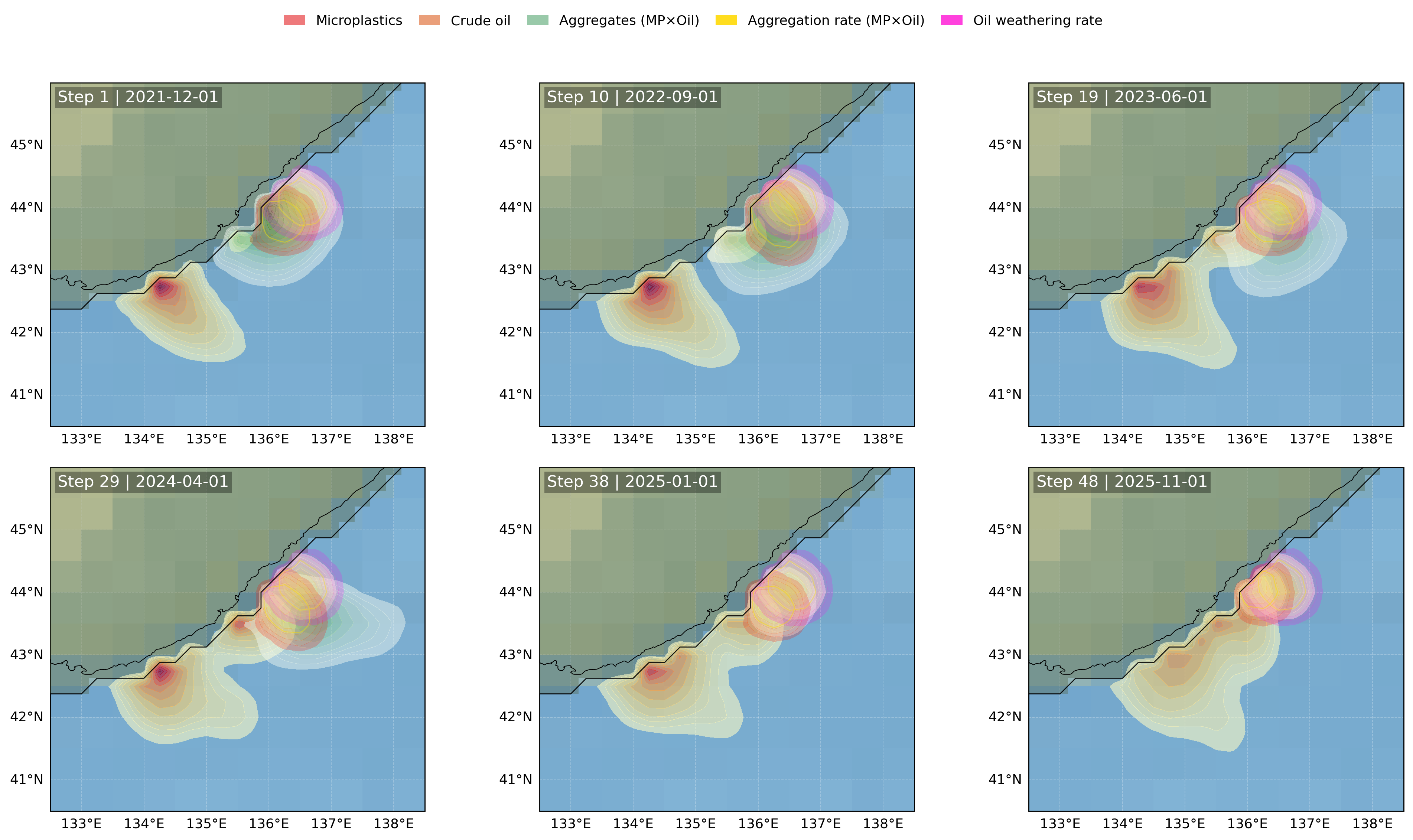}
\caption{Six sampled times from the dataset-driven surface multispecies proxy. The overlays show normalized microplastic, crude-oil, and aggregate fields together with aggregation and oil-weathering rates. The land mask blocks the proxy fields. Because each species is normalized independently after every substep, this panel is qualitative evidence of transport and reaction structure rather than a mass forecast.}
\label{fig:app_multispecies}
\end{figure*}

The supplementary visualization uses a separate two-dimensional proxy driven by a selected depth slice of the dataset currents. For any field $F$, one transport substep is
\begin{equation}
\mathcal T_n(F)=F+\Delta t\left[-u_nF_x-v_nF_y
+\kappa(F_{xx}+F_{yy})\right].
\label{eq:surface_transport}
\end{equation}
Let $A$, $B$, and $C$ denote normalized microplastic, crude-oil, and aggregate fields, respectively. Transport first gives $A^\star=\mathcal T_n(A^n)$, $B^\star=\mathcal T_n(B^n)$, and $C^\star=\mathcal T_n(C^n)$. With $[x]_+=\max(x,0)$, aggregation and weathering are then
\begin{equation}
r_a=k_a[A^\star]_+[B^\star]_+,
\qquad r_w=k_w[B^\star]_+,
\label{eq:surface_reactions}
\end{equation}
and the implemented update before masking and normalization is
\begin{align}
\widetilde A &=[A^\star-\Delta t\,r_a]_+ +0.010S_A, \nonumber\\
\widetilde B &=[B^\star-\Delta t(r_a+r_w)]_+ +0.008S_B, \nonumber\\
\widetilde C &=[C^\star+\Delta t\,r_a]_+ .
\label{eq:surface_multispecies}
\end{align}
Gaussian source maps $S_A$ and $S_B$ continuously replenish the two parent fields. Land-mask cells are then set to zero, and every nonzero species field is independently divided by its spatial maximum. Consequently, Figure~\ref{fig:app_multispecies} shows relative plume shape and reaction overlap rather than calibrated concentration or mass-conserving forecasts. This surface proxy uses centered numerical gradients and must not be conflated with the upwind three-dimensional solver in Equations~\ref{eq:pollution_balance}--\ref{eq:pollution_cfl}.

Building on the distinct localization, containment, and discrete-cleanup semantics defined in the main paper, the exact canonical SPC presets are: 12 sources, 8\,s dwell, and one required agent at medium difficulty; 6 sources, 6\,s dwell, and two required agents at hard difficulty.

\subsection{Optional Robot Operating System 2 (ROS 2) Jazzy bridge}

The bridge maps core $(x,y,z)=$(east, down, north) to the ROS East-North-Up (ENU) frame $(x,z,-y)$. External commands traverse the same clipping, dynamics, current, constraints, recorder, and task-update path as built-in controllers; internal-control mode can publish observations without accepting external motion commands.

\begin{table}[t]
\centering
\scriptsize
\setlength{\tabcolsep}{3pt}
\renewcommand{\arraystretch}{1.08}
\begin{tabularx}{\columnwidth}{l X}
\toprule
\textbf{Interface} & \textbf{ROS type and meaning} \\
\midrule
\texttt{/agent\_NNN/odom} & \texttt{nav\_msgs/Odometry}; ENU pose and velocity \\
\texttt{/agent\_NNN/current} & \texttt{Vector3Stamped}; ENU current \\
\texttt{/agent\_NNN/pollution} & \texttt{Float32}; local scalar probe \\
\texttt{/agent\_NNN/bathymetry} & \texttt{Float32}; positive water depth \\
\texttt{/agent\_NNN/cmd\_vel} & subscribed ENU velocity command \\
\texttt{/clock}, \texttt{/episode/metrics} & simulation time and JavaScript Object Notation episode state \\
\texttt{/reset}, \texttt{/step}, \texttt{/get\_task\_state} & \texttt{std\_srvs/Trigger} services \\
\bottomrule
\end{tabularx}
\caption{ROS 2 interface. Topic prefixes are \texttt{/oneocean/agents}; agent IDs are zero-padded.}
\label{tab:app_ros_contract}
\end{table}

The package was built and launched in a ROS 2 Jazzy container; topic discovery, one externally commanded step, state query, and reset completed successfully. This is an interface smoke test, not a ROS performance result, and ROS remains an optional dependency outside the core requirements.

\section{Metrics and Reproducibility Protocol}
\label{app:protocol}

\begin{table*}[t]
\centering
\scriptsize
\setlength{\tabcolsep}{4pt}
\renewcommand{\arraystretch}{1.08}
\begin{tabularx}{\textwidth}{l l l X}
\toprule
\textbf{Diagnostic} & \textbf{Budget} & \textbf{Configuration} & \textbf{Purpose} \\
\midrule
Planner audit & 3 seeds per task and method & 72 episodes; stride 200; 96-token cap & Validity, fallback, completion, and failure diagnosis. \\
Dynamics diagnostic & 20 episodes per task and difficulty & 10 tasks; medium and hard; three dynamics modes & Sensitivity to engineering dynamics. \\
Current-gain sweep & 5 seeds per task and method & 300 episodes; hard; 10 tasks; $g\in\{0,1,2\}$ & Data-current disturbance sensitivity. \\
Synthetic-tide stress & 10 seeds per task and method & 200 episodes; hard; 10 tasks; 0.45\,m\,s$^{-1}$ amplitude & Periodic-disturbance sensitivity. \\
External pollution diagnostic & 3 episodes per scene and configuration & Four scenes; two currents; two field models & Localization-model comparison. \\
External scaling & 10 episodes per scene and team size & 80 episodes; two scenes; $N\in\{2,4,8,10\}$ & Scene-backed team-size sensitivity. \\
\bottomrule
\end{tabularx}
\caption{Budgets for supplementary diagnostic studies. Headline experiment budgets remain attached to their result tables in the main paper.}
\label{tab:app_protocol_budgets}
\end{table*}

\subsection{Metric definitions}

The main paper defines Success Rate (SR), timeout-aware episode time $T$, and the action-energy proxy $E$. The additional team-level aggregate used in the result tables is the near-collision ratio
\begin{equation}
\mathrm{Coll}=\frac{1}{T_{\mathrm{ep}}}\sum_{t=1}^{T_{\mathrm{ep}}}\mathbf 1\!\left[\min_{i<j}\|p_t^{(i)}-p_t^{(j)}\|_2\le r_{\mathrm{coll}}\right].
\end{equation}
Task-primary metrics follow the per-task contracts in Appendix Sec.~\ref{app:tasks} and are not pooled across incompatible physical meanings.

\subsection{Episode and task aggregation}

For task $k$ with $M_k$ episodes, let $s_{ke}\in\{0,1\}$ indicate success and $H_{ke}$ the timeout in seconds. The reported success rate and timeout-aware duration are
\begin{align}
\mathrm{SR}_k&=\frac{1}{M_k}\sum_e s_{ke},\\
T_k&=\frac{1}{M_k}\sum_e\left[s_{ke}t_{ke}^{\rm succ}
                         +(1-s_{ke})H_{ke}\right],
\end{align}
where $t_{ke}^{\rm succ}$ is the success time for a completed episode. The success-only duration, when reported separately, averages $t^{\rm succ}$ over successful episodes and is undefined if none succeeds.

An episode's action-energy proxy sums over both agents and executed steps:
\begin{equation}
E_{ke}=\Delta t\sum_{n=0}^{L_{ke}-1}\sum_{i=1}^{N}
                   \|\Pi_v(a_{i,n})\|_2^2.
\end{equation}
Here $L_{ke}$ is the executed step count. The recorded current and realized velocity do not enter this command-based quantity.  The near-collision ratio above uses one indicator per step, even if several agent pairs are close in that step; tables expressed in percent multiply this ratio by 100.

Station error is accumulated after each motion update. Writing $\bar d_n=N^{-1}\sum_i\|p_i^n-p_g\|_2$, its episode RMS is $[L^{-1}\sum_n\bar d_n^2]^{1/2}$. The route-tracking diagnostic analogously averages distance to the active waypoint across agents and time. It measures the waypoint-target error associated with the current route stage. For an overall success rate across $K$ tasks, the macro-average is $K^{-1}\sum_k\mathrm{SR}_k$, which gives every task equal weight. With equal episode counts this also equals the pooled episode success rate. Physical task metrics retain their own units and are averaged within their task and configuration.

\subsection{Controller observations and decisions}

The heuristic, BC, and LLM-assisted controllers share the action interpretation and the motion update in Appendix Sec.~\ref{app:models_ros}. Heuristics track task targets, BC maps observations to velocity, and the LLM selects discrete task decisions.

\paragraph{Goal-directed control.}
For an assigned target $q_i$, the proportional controller produces
\begin{equation}
a_i=\Pi_v\!\left(k_p(q_i-p_i)\right).
\end{equation}
The navigation gain is $k_p=0.9$ and the station-keeping gain is $0.8$. Intermediate targets come from the task state: the active waypoint, assigned source, formation offset, or stage of a coordinated task. This controller corrects positional error at each step under the applied current. The pollution-localization heuristic additionally uses the agents' scalar probes to estimate a horizontal concentration gradient and coordinate exploration around the highest observed concentration.

\paragraph{Behavior Cloning.}
The observation for each training example is
\begin{equation}
o_i=[(q_i-p_i)^\top,\ y_i,\ C(p_i),\ c_{i,x},\ c_{i,z},\ e_k^\top]^\top,
\end{equation}
where $e_k$ is a one-hot task indicator. It contains seven continuous features followed by the task encoding. The policy is a multilayer perceptron with two ReLU hidden layers and a three-dimensional linear output. Feature and action means and standard deviations are estimated on the training split. With $\widehat o$ and $\widehat a$ denoting standardized values, the learning objective is
\begin{equation}
\mathcal L_{\rm BC}(\theta)=\frac{1}{3M}\sum_{m=1}^{M}
            \|f_\theta(\widehat o_m)-\widehat a_m\|_2^2.
\end{equation}
AdamW minimizes this objective. At inference the output is transformed back to action units and projected onto the permitted speed ball. The reference training configuration uses hidden widths $(64,64)$, 12 epochs, batch size 4096, learning rate $10^{-3}$, zero weight decay, and a randomly selected 10\% validation split; the trained checkpoint records its configuration and normalization statistics.

Demonstrations align the recorded command with its task target and local observations. The trajectory converter retains steps with a recorded target. For the kinematic recording convention, which stores the post-step position, it reconstructs the corresponding pre-step position as $p_n=p_{n+1}-\Delta t(a_n+c_n)$ before forming the target displacement.

\paragraph{LLM-assisted planning.}
At a planning event, the task state is summarized by the task identifier, step, team size, agent positions, and the relevant source or waypoint state. Coordinates are rounded to two decimal places. The planner returns discrete source assignments or waypoint choices, which are checked for admissible indices and task state before use. The task controller then tracks the resulting targets using the same bounded velocity control. Between planning events the current decisions remain active. Failed or invalid responses invoke the task's deterministic fallback. The audit in Appendix Sec.~\ref{app:core_diagnostics} separates decision validity, fallback frequency, task completion, and inference cost, to characterize discrete planning within the closed-loop episode.

\subsection{Supplementary diagnostic budgets}

All diagnostics reuse the same serialized tier presets, controller settings, dataset indices, and environment provenance as their corresponding headline study; only the diagnostic factors named in Table~\ref{tab:app_protocol_budgets} are varied.

\subsection{Compute and run manifests}

The main paper reports the compute platform. For provenance, each run stores an integer seed, normalized configuration, task and dynamics specification snapshots, environment-product indices, software and Git state, episode metrics, and recorder streams. Aggregation scripts consume these artifacts to regenerate comma-separated-value and Markdown tables rather than transcribing console output.

The reproducibility boundary is explicit: data validation and ROS adapter tests are release tests; they are not added to performance averages. Likewise, the three-seed planner audit and three-episode-per-scene pollution comparison diagnose mechanisms but do not replace the larger headline budgets.

\subsection{Model boundaries and extension priorities}
\label{app:limitations}

The pollution fields are controlled sensing signals rather than in-situ labels or validated spill forecasts. Surface Pollution Cleanup completes discrete sources after a dwell requirement, whereas continuous containment removes mass from a concentration field. Underwater Pollution Lift instead measures coordinated payload lifting; these objectives have distinct completion criteria. Tide attenuation remains an engineering approximation rather than a phase-accurate three-dimensional tide model. The ROS~2 bridge has passed message, reset, frame-conversion, and command smoke tests, but it has not been benchmarked for latency, determinism, middleware load, or hardware-in-the-loop performance. The highest-priority extensions are measured three-dimensional flow, identified vehicle coupling through Computational Fluid Dynamics (CFD) or system identification, physically validated pollutant transport, and ROS-in-the-loop experiments. These additions can preserve the current task and reporting contracts while increasing physical realism.

\section{Additional Core-Backend Diagnostics}
\label{app:core_diagnostics}

\subsection{Dynamics-mode sensitivity}

Table~\ref{tab:app_dynamics_ablation} selects the hard-setting tasks with the largest observed dynamics sensitivity from a 20-episode-per-task diagnostic.

\begin{table}[t]
\centering
\scriptsize
\setlength{\tabcolsep}{4pt}
\renewcommand{\arraystretch}{1.10}
\begin{tabularx}{\columnwidth}{l *{3}{Y}}
\toprule
\textbf{Hard task} & \textbf{Kinematic SR} & \textbf{3DoF SR} & \textbf{6DoF SR} \\
\midrule
Formation Transit & 55\% & 50\% & 50\% \\
Surface Cleanup & 95\% & 85\% & 85\% \\
Pollution Lift & 100\% & 75\% & 75\% \\
All-task macro & 85.5\% & 81.5\% & 81.5\% \\
\bottomrule
\end{tabularx}
\caption{Selected dynamics sensitivity at hard difficulty (20 episodes per task and mode).}
\label{tab:app_dynamics_ablation}
\end{table}

Across all ten hard tasks, macro SR is 85.5\% for kinematic and 81.5\% for both 3DoF and 6DoF. The identical 3DoF and 6DoF SRs under this controller do not establish physical equivalence; they show that bounded roll and pitch did not change terminal success in this diagnostic, whereas replacing the velocity state with direct kinematic integration did.

\subsection{Current and synthetic-tide stress}

\begin{table}[!t]
\centering
\scriptsize
\setlength{\tabcolsep}{2pt}
\renewcommand{\arraystretch}{1.08}
\begin{tabularx}{\columnwidth}{@{}>{\raggedright\arraybackslash}X l r r r r@{}}
\toprule
\textbf{Stress} & \textbf{Method} & \textbf{$N$} & \textbf{SR} & \textbf{$E$} & \textbf{Coll.} \\
\midrule
Data current, $g=0$ & Heuristic & 50 & 48.0 & 3232.5 & 5.80 \\
Data current, $g=1$ & Heuristic & 50 & 20.0 & 3371.6 & 8.24 \\
Data current, $g=2$ & Heuristic & 50 & 8.0 & 3775.0 & 9.86 \\
Data current, $g=0$ & BC & 50 & 6.0 & 1546.8 & 4.14 \\
Data current, $g=1$ & BC & 50 & 12.0 & 1807.1 & 4.96 \\
Data current, $g=2$ & BC & 50 & 4.0 & 2323.3 & 2.44 \\
Synthetic tide & Heuristic & 100 & 17.0 & 3489.2 & 6.29 \\
Synthetic tide & BC & 100 & 1.0 & 1557.6 & 6.49 \\
\bottomrule
\end{tabularx}
\caption{Archived hard-setting disturbance diagnostics across all ten tasks. SR and Coll. are percentages. Each data-current row contains five episodes per task. The synthetic-tide rows contain ten episodes per task, use $g=0$, and apply a 0.45\,m\,s$^{-1}$ sinusoid with a 300\,s period. These focused sweeps diagnose sensitivity and are separate from the headline budgets.}
\label{tab:app_disturbance_stress}
\end{table}

Heuristic SR decreases from 48.0\% without data-current forcing to 8.0\% at doubled current, while its energy and collision exposure increase. BC is not monotonic at the smaller gains, but reaches its lowest SR at $g=2$. The synthetic-tide stress also separates the methods, with 17.0\% SR for the heuristic and 1.0\% for BC. Because the current-gain cells contain only five episodes per task, these results establish a sensitivity trend rather than precise task-level rankings.

\subsection{Planner validity, fallback, and cost}

The instrumented audit contains three medium-difficulty seeds for each of eight methods and three tasks, for 72 episodes total. Tables~\ref{tab:app_llm_model_outcomes} and \ref{tab:app_llm_model_costs} aggregate each Large Language Model (LLM) across its nine episodes. Valid-plan ratio is weighted by planner calls; latency and tokens are weighted by uncached calls.

\begin{table}[!ht]
\centering
\scriptsize
\setlength{\tabcolsep}{2.5pt}
\renewcommand{\arraystretch}{1.08}
\begin{tabularx}{\columnwidth}{l *{5}{Y}}
\toprule
\textbf{LLM} & \textbf{SR} & \textbf{Calls} & \textbf{Valid} & \textbf{Fallback} & \textbf{Comp.} \\
\midrule
ChatGLM3-6B & 0.556 & 49 & 0.367 & 0.633 & 0.758 \\
Llama2-7B & 0.333 & 60 & 0.950 & 0.050 & 0.736 \\
Llama3-8B & 0.444 & 61 & 0.869 & 0.131 & 0.786 \\
Mistral-7B & 0.556 & 55 & 0.800 & 0.200 & 0.804 \\
Qwen2-7B & 0.556 & 43 & 0.814 & 0.186 & 0.790 \\
Qwen2.5-7B & 0.667 & 59 & 0.864 & 0.136 & 0.809 \\
\bottomrule
\end{tabularx}
\caption{Per-model planning outcomes across SPC, AS, and PI (nine episodes per LLM).}
\label{tab:app_llm_model_outcomes}
\end{table}

\begin{table}[!ht]
\centering
\scriptsize
\setlength{\tabcolsep}{3pt}
\renewcommand{\arraystretch}{1.08}
\begin{tabularx}{\columnwidth}{l *{3}{Y}}
\toprule
\textbf{LLM} & \makecell{\textbf{Latency}\\\textbf{ms per call}} & \makecell{\textbf{Prompt}\\\textbf{tokens per call}} & \makecell{\textbf{Output}\\\textbf{tokens per call}} \\
\midrule
ChatGLM3-6B & 1466.1 & 628.1 & 76.7 \\
Llama2-7B & 7429.6 & 633.7 & 728.7 \\
Llama3-8B & 2068.1 & 454.1 & 475.9 \\
Mistral-7B & 5150.6 & 616.5 & 676.7 \\
Qwen2-7B & 2370.6 & 603.3 & 628.6 \\
Qwen2.5-7B & 2184.5 & 600.6 & 623.9 \\
\bottomrule
\end{tabularx}
\caption{Uncached per-call planning cost in the same audit.}
\label{tab:app_llm_model_costs}
\end{table}

\begin{table}[t]
\centering
\scriptsize
\setlength{\tabcolsep}{3pt}
\renewcommand{\arraystretch}{1.08}
\begin{tabularx}{\columnwidth}{l *{5}{Y}}
\toprule
\textbf{Task} & \textbf{SR} & \textbf{Valid} & \textbf{Fallback} & \makecell{\textbf{Latency}\\\textbf{ms per call}} & \textbf{Completion} \\
\midrule
SPC & 0.833 & 0.430 & 0.570 & 3199.0 & 0.972 \\
AS & 0.278 & 0.831 & 0.169 & 3602.0 & 0.758 \\
PI & 0.444 & 0.891 & 0.109 & 3337.1 & 0.611 \\
\bottomrule
\end{tabularx}
\caption{Cross-model planner diagnostics by task (six LLMs, three seeds each).}
\label{tab:app_llm_task_audit}
\end{table}

Model-level and task-level rankings also diverge: Llama2-7B attains 0.950 validity but only 0.333 SR, whereas ChatGLM3-6B attains 0.367 validity and 0.556 SR. At task level, cleanup's low valid ratio coexists with high completion because invalid or duplicate assignments can fall back to a competent deterministic allocator; AS can revisit covered cells despite valid syntax, and PI still requires physical leak encounters between low-frequency plans. Planner validity is therefore a mechanism diagnostic, not a substitute for task success.

\section{Additional External-Backend Diagnostics}
\label{app:external_diagnostics}

The external pollution study compares an analytic advected Gaussian probe with OCPNet-3D in Dam, OpenWater, PierHarbor, and SimpleUnderwater. Each combination of scene, model, and current setting uses three localization episodes; Table~\ref{tab:app_pollution_localization} therefore aggregates 12 episodes per row. It is a focused observation-model diagnostic rather than a high-confidence performance ranking.

\begin{table}[t]
\centering
\scriptsize
\setlength{\tabcolsep}{2pt}
\renewcommand{\arraystretch}{1.10}
\begin{tabularx}{\columnwidth}{@{}l l *{4}{Y}@{}}
\toprule
\textbf{Scale} & \textbf{Probe} & \textbf{$N$} & \makecell{\textbf{SR}\\\textbf{(\%)}} & \makecell{\textbf{Error}\\\textbf{(m)}} & \makecell{\textbf{$\Delta$ error}\\\textbf{(m)}} \\
\midrule
1.0 & Analytic & 12 & 41.7 & 25.93 & 0.00 \\
1.0 & OCPNet-3D & 12 & 50.0 & 20.81 & $-5.12$ \\
2.5 & Analytic & 12 & 50.0 & 26.00 & 0.00 \\
2.5 & OCPNet-3D & 12 & 50.0 & 22.64 & $-3.36$ \\
\bottomrule
\end{tabularx}
\caption{Pollution-localization diagnostic over four scenes (12 episodes per row). $\Delta$ error is measured against the analytic probe at the same current scale.}
\label{tab:app_pollution_localization}
\end{table}

OCPNet-3D lowers mean localization error by 5.12\,m at nominal current and 3.36\,m under stronger current, while matching or improving aggregate success. This result supports the narrower claim that richer regeneratable concentration structure can change sensing difficulty; it does not validate OCPNet as a real spill forecast. Containment outputs are not collapsed into the same table because analytic remaining-mass state and OCPNet grid mass have different scales and should not be compared as a shared physical quantity.

\subsection{Scene-backed cleanup scaling}

\begin{table}[!ht]
\centering
\scriptsize
\setlength{\tabcolsep}{5pt}
\renewcommand{\arraystretch}{1.08}
\begin{tabular}{r r r r}
\toprule
\textbf{Agents} & \textbf{Episodes} & \textbf{SR (\%)} & \makecell{\textbf{Removed mass}\\\textbf{(kg)}} \\
\midrule
2 & 20 & 40.0 & 0.2081 \\
4 & 20 & 100.0 & 0.4302 \\
8 & 20 & 100.0 & 0.4817 \\
10 & 20 & 100.0 & 0.4886 \\
\bottomrule
\end{tabular}
\caption{Analytic-containment scaling at nominal current, aggregated over OpenWater and PierHarbor with ten episodes per scene and team size.}
\label{tab:app_external_scaling}
\end{table}

Increasing the team from two to four agents raises aggregate success from 40\% to 100\% and approximately doubles removed mass. The smaller gains from four to ten agents indicate saturation under the fixed 15\,s containment horizon. Removed mass is comparable within this analytic-field sweep only and is not equated with OCPNet grid mass.

The route-transfer results retained in the main paper reveal substantial scene dependence despite a shared task contract. Together, the route and pollution diagnostics motivate reporting scene, current scale, dataset clock, controller profile, and observation model in every external-backend manifest rather than treating ``external simulator'' as a single condition.

\end{document}